\documentclass[journal]{IEEEtran}
\ifCLASSINFOpdf
\else
\fi

\usepackage[acronym]{glossaries}
\newacronym{cnn}{CNN}{Convolutional Neural Network}
\newacronym{dbnn}{DBNN}{Decision Based Neural Network}
\newacronym{rf}{RF}{Random Forest}
\newacronym{knn}{k-NN}{k-Nearest Neighbour}
\newacronym{pca}{PCA}{Principle Component Analysis}
\newacronym{ar}{AR}{AutoRegressive}
\newacronym{svm}{SVM}{Support Vectors Machine}
\newacronym{rnn}{RNN}{Recurrent Neural Network}
\newacronym{lstm}{LSTM}{Long Short-Term Memory}
\newacronym{gru}{GRU}{Gated Recurrent Unit}
\newacronym{nn}{NN}{Neural Network}
\newacronym{fcl}{FCL}{Fully-Connected Layer}
\newacronym{relu}{ReLU}{Rectified Linear Unit}
\newacronym{ldac}{LDAC}{Linear Discriminant Analysis Classifier}
\newacronym{mlp}{MLP}{Multilayer Perceptron}
\newacronym{bn}{BN}{Bayes Network}
\newacronym{rbf}{RBF}{Radial Basis Function}
\newacronym{mle}{MLE}{Maximum Likelihood Estimation}
\newacronym{sodp}{SODP}{Second Order Difference Plot}
\newacronym{ncn}{NCN}{Normalize-Convolute-Normalize}
\newacronym{lga}{LGA}{Logarithmic Grid Analysis}
\newacronym{lda}{LDA}{Linear Discriminant Analysis}
\newacronym{arm}{ARM}{AutoRegressive Model}
\newacronym{svd}{SVD}{Singular Value Decomposition}
\newacronym{cce}{CCE}{Categorical Cross-Entropy}
\newacronym{bce}{BCE}{Binary Cross-Entropy}
\newacronym{dl}{DL}{Deep Learning}
\newacronym{ml}{ML}{Machine Learning}
\newacronym{ica}{ICA}{Independent Component Analysis}
\newacronym{tm}{TM}{Template Matching}

\newacronym{eer}{EER}{Equal Error Rate}
\newacronym{far}{FAR}{False Acceptance Rate}
\newacronym{frr}{FRR}{False Rejection Rate}
\newacronym{tp}{TP}{True Positive}
\newacronym{tn}{TN}{True Negative}
\newacronym{fp}{FP}{False Positive}
\newacronym{fn}{FN}{False Negative}
\newacronym{bpf}{bpf}{beats per frame}
\newacronym{se}{Se}{Sensitivity}
\newacronym{sp}{Sp}{Specificity}
\newacronym{ppv}{PPV}{Positive Predictive Value}
\newacronym{ir}{IR}{Identification Rate}
\newacronym{fpir}{FPIR}{False-Positive Identification-error Rate}
\newacronym{roc}{ROC}{Receiver Operating Characteristic}
\newacronym{cer}{CER}{Crossover Error Rate}
\newacronym{cm}{CM}{Confusion Matrix}

\newacronym{nsrdb}{NSRDB}{Normal Sinus Rhythm Database}
\newacronym{ptbdb}{PTBDB}{Physikalisch-Technische Bundesanstalt}
\newacronym{ltst}{LTSTDB}{Long Term ST Database}
\newacronym{butqdb}{BUTQDB}{Brno University of Technology ECG Quality Database}
\newacronym{mitdb}{MIT-BIHDB}{MIT-BIH Arrhythmia Database}
\newacronym{svdb}{SVDB}{MIT-BIH supraventricular arrhythmia database}
\newacronym{disciri}{DiSciRiDB}{Charles Sturt Diabates Complication Screening Initiative}
\newacronym{gudb}{GUDB}{Glasgow University Database}
\newacronym{cybhi}{CYBHi}{Check Your Biosignals Here Initiative}
\newacronym{uoftdb}{UofTDB}{University of Toronto Database}

\newacronym{ecg}{ECG}{Electrocardiogram}
\newacronym{ekg}{EKG}{Elektrokardiogramm}
\newacronym{ecm}{ECM}{Electrocardiomatrix}
\newacronym{ekm}{EKM}{Elektrokardiomatrix}
 \newacronym{eeg}{EEG}{Electroencephalogram}
\newacronym{see}{SEE}{Shannon Energy Envelope}
\newacronym{rle}{RLE}{Run-Length Encoding}
\newacronym{hcw}{HCW}{Hybrid Complex Wavelet}
\newacronym{hrv}{HRV}{Heart Rate Variability}
\newacronym{pt}{PT}{Pan and Tompkins}
\newacronym{eda}{EDA}{Electrodermal Activity}
\newacronym{emg}{EMG}{Electromyography}
\newacronym{lca}{LCA}{Low Cardiovascular Activity}
\newacronym{hca}{HCA}{High Cardiovascular Activity}

\newacronym{cvd}{CVD}{Cardiovascular Disease}
\newacronym{afib}{AFIB}{Atrial Fibrillation}
\newacronym{afl}{AFL}{Atrial Flutter}
\newacronym{chf}{CHF}{Congestive Heart Failure}
\newacronym{veb}{VEB}{Ventricular Ectopic Beats}
\newacronym{sveb}{SVEB}{Supraventricular Ectopic Beats}

\newacronym{gpu}{GPU}{Graphical Process Unit}
\newacronym{ovr}{OvR}{One-vs-the-rest}
\newacronym{elektra}{ELEKTRA}{ELEKTRokardiomatrix Application to biometric identification with Convolutional Neural Networks}
\newacronym{ppg}{PPG}{Photoplethysmogram}
\newacronym{rfid}{RFID}{Radio Frequency Identification}
\newacronym{dna}{DNA}{Deoxyribonucleic acid}
\makeglossaries 
\usepackage{booktabs}
\usepackage{pifont}
\usepackage{caption}
\usepackage{multirow}
\usepackage{url}
\usepackage{subcaption}
\usepackage{algorithm}
\usepackage{color,soul}
\newcommand{\cmark}{\ding{51}} 
\newcommand{\xmark}{\ding{55}} 
\usepackage[square,sort,comma,numbers]{natbib}
\usepackage{graphicx}
\usepackage{algorithmic}

\begin{document}
%
\title{ECG-Based Patient Identification: A Comprehensive Evaluation Across Health and Activity Conditions}
%
%
%

\author{\IEEEauthorblockN{Caterina Fuster-Barceló\IEEEauthorrefmark{1}\IEEEauthorrefmark{2},
Carmen Cámara\IEEEauthorrefmark{3} and Pedro Peris-López\IEEEauthorrefmark{3}}\\
\IEEEauthorrefmark{1}Bioengineering Department, Universidad Carlos III de Madrid, Spain,
\IEEEauthorrefmark{2}Bioengineering Division, Instituto de Investigación Sanitaria Gregorio Marañón, Madrid, Spain,
\IEEEauthorrefmark{3}Computer Science and Engineering Department, Universidad Carlos III de Madrid, Spain}

%
%

\markboth{Preprint, 2024}%
{C. Fuster-Barceló \MakeLowercase{\textit{et al.}}: ECG-Based Patient Identification: A Comprehensive Evaluation Across Health and Activity Conditions}
%



\maketitle

\begin{abstract}
Over the course of the past two decades, a substantial body of research has substantiated the viability of utilising cardiac signals as a biometric modality. This paper presents a novel approach for patient identification in healthcare systems using electrocardiogram signals.  A convolutional neural network (CNN) is employed to classify users based on electrocardiomatrices, a specific type of image derived from ECG signals. The proposed identification system is evaluated in multiple databases, {achieving up to 99.84\% accuracy on healthy subjects, 97.09\% on patients with cardiovascular diseases, and 97.89\% on mixed populations including both healthy and arrhythmic patients. The system also performs robustly under varying activity conditions, achieving 91.32\% accuracy in scenarios involving different physical activities. These consistent and reliable results, with low error rates such as a FAR of 0.01\% and FRR of 0.157\% in the best cases, demonstrate the method's significant advancement in subject identification within healthcare systems. By considering patients' cardiovascular conditions and activity levels, the proposed approach addresses gaps in the existing literature, positioning it as a strong candidate for practical applications in real-world healthcare settings.}
\end{abstract}


%
\IEEEpeerreviewmaketitle

\section{Introduction}
Identifying patients is a crucial aspect of providing quality healthcare \citep{alkhaqani2023patient}. In the event of critically ill, elderly, or disabled patients who require frequent medical treatments, rapid and easy identification is vital~\citep{nigam2022biometric}. Patient misidentification is one of the leading causes of medical errors and medical malpractice in hospitals and has been recognised as a serious risk to patient safety~\citep{aguilar2006positive}. {This highlights a significant challenge in healthcare environments, where accurate and efficient patient identification is essential to avoid serious errors. The aim of this study is to address these identification challenges by proposing a novel approach that improves accuracy and efficiency in patient identification, while also being inclusive of patients with CVD and those engaged in varying activity levels.} 

{Current identification methods, such as wristbands, though commonly used, face significant limitations. }Patient identification problems, such as the use of multiple names and identities or the lack of identification documents when patients are non-residents, have been reported~\citep{mccoy2013matching, riplinger2020patient}. This can result in errors such as the administration of the wrong medication to the wrong patient, incorrect diagnosis, inappropriate treatment, delays, and cancellation of operations, among others. For example, a 2016 study classified 7,600 out of 10,915 events as wrong-patient events, involving issues related to patient identification, such as patient misidentification and duplicate records~\citep{patientid_executivesummary}. 
In a survey performed by Patient Now Organisation in 2022, it is stated that, on average, organisations report spending 109.6 hours per week resolving patient identity issues and spend \$1.3M annually on patient resolution~\citep{patientnow}.
In addition, patient identification can cause harm to patients, as well as economic costs to the health system, with up to 10-20\% errors from patient misidentification resulting in harm to patients~\citep{lippi2017managing}. {Furthermore, } there are concerns related to data privacy and security, as patient information must be protected against unauthorised access and breaches~\citep{hathaliya2020securing}. 

Wristbands, a type of identification band worn on the wrist, are used in healthcare facilities to identify patients and ensure that proper treatment and care are received. Typically, wristbands include the patient's name, identification number, and other pertinent information, such as allergies, medical conditions, and medications. A major issue with wristbands is that the information contained in each wristband is not standardised and is often entered manually, which can lead to misidentification of patients~\citep{hemesath2015educational, tase2015patient}. The loss of wristbands is also a concern. In a study conducted in 712 hospitals in the USA, 2,463,727 wristbands were examined, and 2.7\% (67,289) of them had errors, with 49.5\% of these errors due to the absence of wristbands~\citep{renner1993wristband}. A similar study was also conducted in 204 small hospitals, where 451,436 identification wristbands were examined and 28,800 (5.7\%) had errors, with the majority (64.4\%) of these errors being related to the absence of wristbands~\citep{de2019interventions}. As stated above, losing a wristband can lead to misidentification of the patient and incorrect prescriptions for medications. Despite these limitations, wristbands are still used in the healthcare system. 

To mitigate these risks, some hospitals and heritage sites may have protocols in place to ensure that wristbands are properly secured and replaced if lost. They may also use other forms of identification, such as biometric tokens, \gls{rfid} tags or smart cards, to supplement or replace wristbands~\citep{aguilar2006positive}. User identification technologies are advantageous because, despite other identification systems, biometric data is more difficult to ``steal, exchange or forget''~\citep{riplinger2020patient}. However, existing biometric methods, such as ECG-based identification, often fail to consider the distinct impacts of cardiovascular diseases (CVD) and varying activity conditions on identification accuracy, representing a critical gap in the literature.

\begin{figure}
    \centering
    \includegraphics[width=0.45\textwidth]{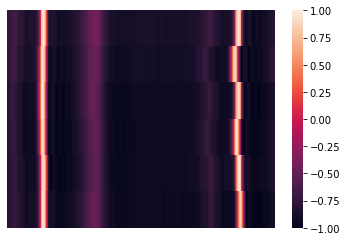}
    \caption{Elektrokardiomatrix or EKM, image used to perform patient identification} 
    \label{fig:ekm}
\end{figure}

In response to these challenges, we propose a novel approach using ECG signals for patient identification that uniquely addresses these gaps by analysing the effects of CVD and activity levels separately. This method refines and implements the approach outlined in~\citep{ELEKTRA2022} and applies it to to various databases, comprising a diverse range of users, including those with various \gls{cvd} and those participating in various activities. The identification method involves segmenting the \gls{ecg} signal into windows of peaks (heartbeats), aligning each peak (beat) to construct a matrix, and transforming the matrix into a heatmap (resulting in an image called \gls{ekm} seen in Fig. \ref{fig:ekm}), which serves as the image input for the patient identification system. A convolutional neural network is then utilised to classify patients based on these images constructed from ECG signals. The conversion of the \gls{ecg} signal to a heatmap, referred to as the \gls{ekm}, is introduced for the purpose of patient identification for the first time. The application of this methodology for the identification, classification, or diagnosis of \gls{cvd} has been previously documented in the literature (cited in references~\citep{ecm, XU2018955, STROKEAHA025361, 9344257, salinas2021detection}).

The novelty of this study lies in its distinct focus on evaluating identification performance across different health conditions and activities, a consideration largely overlooked in prior research. The proposed identification system is extensively tested on multiple databases, providing a comprehensive understanding of its potential in real-world scenarios. This study also demonstrates the potential for this method to be implemented in actual healthcare settings, providing a reliable and efficient solution for patient identification.

\textbf{\textit{Contributions.}}
    \begin{enumerate}
        \item Presenting a new method for identifying patients within healthcare systems utilizing Electrocardiogram (ECG) signals.
        \item Refining the approach presented in~\citep{ELEKTRA2022} to facilitate its integration into healthcare systems and facilities.
        \item Extensive testing of the proposed patient identification system on multiple databases, encompassing healthy and unhealthy users, as well as subjects under activity conditions, provides a comprehensive understanding of its potential in real-world scenarios (\gls{nsrdb}, \gls{mitdb}, \gls{ptbdb}, and \gls{gudb}). 
        \item Exemplifying the feasibility of implementing this approach in real-world healthcare settings, thus offering a reliable and efficient solution for patient identification.
    \end{enumerate}

\section{Literature Review}\label{sec:litrev}

The identification of patients in healthcare settings relies on various biometric systems, each with its own set of advantages and limitations. This section provides a comparative analysis of different biometric traits, such as fingerprint, facial, and voice recognition, alongside more advanced methods like ECG and DNA testing, highlighting their applicability, reliability, and inclusivity in patient identification (Table\mbox{~\ref{tab:biom_comparison}}). Additionally, we review existing studies that explore patient identification under different activity conditions, emphasizing the unique challenges and gaps in identifying individuals with varying heart rates and physical states. Finally, we discuss the literature focused on the identification of patients with \gls{cvd}, underscoring the necessity for inclusive and robust identification methods that can accommodate diverse patient conditions.

Biometric methods such as fingerprint recognition~\citep{jaafa2021implementation}, facial recognition~\citep{saleem2023face, 10042627}, palm recognition{\mbox{~\citep{khatun2022comparison, amrouni2023palmprint}}}, \gls{ecg} key generation~\citep{ostad2019robust}, iris recognition~\citep{anne2020feasibility}, {\mbox{\citep{nguyen2024deep}}} and \gls{eeg} identification~\citep{do2019eeg} can be used to supplement or replace wristbands for patient identification in healthcare~\citep{riplinger2020patient}. 

Other biometric systems, such as those based on a \gls{dna} test, may also be regarded as relatively stable\citep{9035151, jacob2021biometric}. It has been proven that \gls{dna} is an effective way to perform user identification with high accuracy rates. In fact, the requisite analysis for a \gls{dna} test is already conducted within healthcare facilities. 

These methods use unique physical characteristics of the patient to identify them and can be integrated into electronic medical record systems to improve patient identification accuracy and reduce the risk of errors in medical treatment. Unfortunately, few approaches for patient identification are currently found within the literature or being utilised within healthcare facilities.

Additionally, smart cards with chips embedded with patient's personal information can also be used to replace wristbands to reduce the risk of lost or stolen wristbands~\citep{aubert2001adoption}. However, the implementation of these methods also raises concerns around data privacy and security.
\begin{table*}[]
    \centering
    \small
    \begin{tabular}{c c c c c c}
    \toprule
         Biometric System & \textbf{User Friendly} & \textbf{Reliability} & \textbf{Inclusivity} & \textbf{Availability} & \textbf{No Awareness} \\
    \midrule
        \textbf{Fingerprint R.} & \cmark & \xmark & \xmark & \xmark & \cmark \\
        \textbf{DNA Test} & \xmark & \cmark & \cmark & \cmark & \cmark \\
        \textbf{Facial R.} & \cmark & \xmark & \cmark & \xmark & \cmark \\
        \textbf{Retina Scan} & \xmark & \cmark & \cmark & \xmark & \xmark \\
        \textbf{Voice R.} & \cmark & \xmark & \xmark & \xmark & \xmark \\
        \textbf{Electrocardiograms} & \cmark & \cmark & \cmark & \cmark & \cmark \\
    \bottomrule
    \end{tabular}
    \caption{Biometric systems for patient identification in healthcare systems and facilities: limitations and advantages.}
    \label{tab:biom_comparison}
    
\end{table*}

There exist some limitations to the methods currently employed for patient identification in healthcare systems and facilities. As shown in Table \ref{tab:biom_comparison}, a comparison is provided between these existing biometric methods applied to patient identification, with respect to various aspects of each method. Regarding fingerprint recognition, which has been extensively researched for patient identification, as demonstrated in~\citep{sohn2020clinical}, {\mbox{\citep{raghavan2024enhanced}}}, there are numerous limitations related to its inclusion. Adermatoglyphia, which is defined as the congenital or acquired loss of the epidermal ridges of the fingertips, commonly known as fingerprints, is one of such limitations. There are various causes of adermatoglyphia such as Kindler syndrome, chronic hand eczema, peeling skin syndrome, and others~\citep{batool2022causes}. Additionally, research conducted in~\citep{deneken2022capecitabine} presented a case report showing that capecitabine, a fluoropyrimidine used as a treatment for tumours, has occasionally been reported to cause adermatoglyphia as a secondary effect upon its use. Therefore, it is concluded that the fingerprint is not inclusive enough to be included as an identification method for patients in healthcare systems and facilities.

Similar limitations exist when voice recognition is used as a biometric trait for patient identification. It is not as inclusive as other biometric traits, such as DNA or a facial scan, as there are individuals who are non-verbal or have a communication disorder, which can impede the effectiveness of voice recognition. Additionally, the individual must be awake and aware for the voice recognition to be successful, whereas, with other biometric traits, such as a facial scan, identification can be achieved even when the individual is unconscious. For example, in the case of a retina scan, the patient doesn't need to be awake, but it may not be as user-friendly as other biometric tests, as the eye must be manually opened to perform the test. Additionally, even if the patient is awake, the test may be uncomfortable to perform.
Regarding the reliability of biometric systems, facial recognition issues may arise when applied to patients in a healthcare facility. Factors such as illumination, {environment setup,} pose variation, expression changes, {occlusion}, or even facial paralysis following a stroke may impede the ability to conduct a facial scan~\citep{adjabi2020past, ali2021classical}.
Additional challenges are present with other biometric systems. For example, using DNA testing, which is one of the most inclusive biological traits, for user identification in a patient, requires a minimum of 24 hours in a hospital if laboratories are not busy. Therefore, it may not be feasible to identify the patient immediately.

{A primary focus of this research is the ability to identify patients under varying conditions, including different heart rates ranging from high to low. To the best of our knowledge, this is the first study to evaluate the performance of an identification method on individuals with varying heart rates categorized by specific activities. While some studies have tested identification methods on databases involving subjects in motion, achieving high performance, they do not differentiate between specific activities and heart rates, often combining all heart rate conditions in the same experiments \mbox{\citep{dargie2024identification, 9231814}}. Other works address challenges related to QRS detection from signals recorded during exercise and analyze how activity impacts detection accuracy; however, these studies focus on QRS detection rather than on the identification of individuals \mbox{\citep{apandi2022qrs}}.}

{A comprehensive review of the literature on user identification methods, detailed in the Appendix, reveals that most existing studies perform identification on either healthy users or a mixed pool of healthy users and patients with {\gls{cvd}}. However, none of the reviewed articles specifically differentiate or analyze how {\gls{cvd}} affects the identification process independently. This lack of focus on exclusively studying {\gls{cvd}}-affected populations presents a significant gap in the current literature, as no studies provide distinct identification rates specifically for users with {\gls{cvd}}. Consequently, a direct comparison between our approach and existing methods is challenging since the available studies do not isolate or evaluate the impact of {\gls{cvd}} on identification performance. Our work addresses this critical gap by specifically analyzing and highlighting the influence of {\gls{cvd}} on patient identification, thereby providing insights that are absent in the existing literature.}

\section{Materials and Methods}

\subsection{Data}\label{subsec:data}

The experiments detailed in this article were conducted over four distinct databases, as delineated in Table \ref{tab:ddbb}, each of which was chosen based on its particular features. Firstly, to evaluate the effectiveness of our approach and establish a baseline for an ideal scenario, we conducted experimentation on a database featuring only control users(\gls{nsrdb}). Subsequently, our approach was tested across two databases comprising users with varying \gls{cvd} to more closely reflect real-life healthcare facility scenarios where patients may present with a range of cardiac conditions. The first of these databases, namely the \gls{mitdb}, was chosen due to its prominence in current literature pertaining to cardiovascular disease identification and user identification. The\gls{ptbdb} was also selected, as it includes a range of \gls{cvd} that can be included or excluded depending on the experiment. Finally, the \gls{gudb} was included, providing records from 25 users in diverse situations and allowing our approach to be tested under various heart rates. We consider these four databases to be sufficient to test our approach and obtain a preliminary understanding of its behaviour as a patient identification technique in various perspectives.

The first database used in this study is \gls{nsrdb} (see Table \ref{tab:ddbb}), which is publicly available on Physionet~\citep{nsrdb}. The database comprises 18 healthy subjects (without significant arrhythmias), and records were obtained in the Arrhythmia Laboratory at Beth Israel Hospital in Boston. One record per user is present. The distinctive feature of this database is that it comprises very large records, thereby affording a significant amount of data for testing the presented approach.

The second database selected is widely used \gls{mitdb}, which is also publicly available on Physionet~\citep{mitdb}. This database consists of 48 recordings from 47 different subjects. From the pool of 47 subjects, there are 23 control users and 24 patients who are considered to have significant arrhythmias. In particular, this database combines healthy individuals and patients with \gls{cvd}, mimicking real-world situations.

The \gls{ptbdb} is another publicly available database from Physionet, which comprises 549 records from 290 subjects~\citep{ptbdb}. This database is notable for its inclusion of both patients with various cardiac conditions and healthy subjects. Specifically, the original database contains 52 healthy control users and the remaining individuals with different \gls{cvd}. This database is an unbalanced dataset, with some users having more than one \gls{ekg} recorded. Therefore, only one \gls{ekg} per user was considered in all experiments. The first experiment with this database will examine the entire dataset, while the second experiment will focus on specific \gls{cvd} without including healthy subjects to demonstrate the performance of the presented approach on users with these cardiac conditions. The \gls{cvd} being studied in these experiments are bundle branch block, cardiomyopathy, dysrhythmia, myocardial infarction, myocarditis, and valve heart disease. As such, two approaches will be taken concerning this database: processing the entire dataset and processing only the subjects with the aforementioned \gls{cvd}.

\begin{table*}[t]
\renewcommand{\arraystretch}{1.4}
\centering
\small
\begin{tabular}{c|ccccc|c|c|}
\cline{2-8}
 &
  \multicolumn{5}{c|}{ \textbf{Subject Information}} &
    &
    \\ \cline{2-6}
 &
  \multicolumn{1}{c|}{ \textbf{Number}} &
  \multicolumn{1}{c|}{ \textbf{Male}} &
  \multicolumn{1}{c|}{ \textbf{Women}} &
  \multicolumn{1}{c|}{ \textbf{Ages}} &
   \textbf{Pathologies} &
  \multirow{-2}{*}{ \textbf{\begin{tabular}[c]{@{}c@{}}Number of\\ records\end{tabular}}} &
  \multirow{-2}{*}{ \textbf{\begin{tabular}[c]{@{}c@{}}Sampling\\ frequency (Hz)\end{tabular}}} \\\hline
\multicolumn{1}{|c|}{\textbf{NSRDB}} &
  \multicolumn{1}{c|}{18} &
  \multicolumn{1}{c|}{5} &
  \multicolumn{1}{c|}{13} &
  \multicolumn{1}{c|}{20-50} &
  Healthy Subjects &
  18 &
  1000 \\\hline
\multicolumn{1}{|c|}{\textbf{MITDB}} &
  \multicolumn{1}{c|}{47} &
  \multicolumn{1}{c|}{26} &
  \multicolumn{1}{c|}{22} &
  \multicolumn{1}{c|}{23-89} &
  \begin{tabular}[c]{@{}c@{}}23 random subjects\\ and 24 patients with \\ significant arrhythmias\end{tabular} &
  48 &
  360 \\\hline
\multicolumn{1}{|c|}{\textbf{PTBDB}} &
  \multicolumn{1}{c|}{290} &
  \multicolumn{1}{c|}{209} &
  \multicolumn{1}{c|}{81} &
  \multicolumn{1}{c|}{17-87} &
  \begin{tabular}[c]{@{}c@{}}Different Cardiac\\ Conditions\end{tabular} &
  549 &
  1000 \\ \hline
\multicolumn{1}{|c|}{\textbf{GUDB}} &
  \multicolumn{1}{c|}{25} &
  \multicolumn{1}{c|}{9} &
  \multicolumn{1}{c|}{15} &
  \multicolumn{1}{c|}{18-X} &
  Healthy Subjects &
  100 &
  250 \\\hline
\end{tabular}%
\caption{Database information and characteristics.}
\label{tab:ddbb}
\end{table*}

Inclusion of the \gls{gudb} database among those studied was deemed necessary. This database, which is available through a request to Glasgow University~\citep{gudb}, possesses a unique attribute in that all 25 users were recorded in five different scenarios: sitting, completing a maths test on a tablet, walking on a treadmill, running on a treadmill, and using a hand bicycle. As a result, each user will have five \gls{ecg} recordings. All participants in this database are considered healthy subjects, with no significant \gls{cvd} or health issues. Through this database, an investigation of the identification of patients with varying heartbeat rhythms is proposed.

\subsection{Patient identification}
The methodology developed in~\citep{ELEKTRA2022} (sketched in Figure~\ref{fig:pipeline}) has been implemented for patient identification. This methodology involves the conversion of \gls{ecg}s into heatmaps of aligned R-peaks, which are then used to form a matrix called a \gls{ekm}. The \gls{ekm} is then plotted into a heatmap. A \gls{cnn} architecture, which is both simple and effective, is utilised for patient identification and has been found to offer high accuracy and low error rates.

\subsubsection{Creation of the EKM}
\begin{figure*}
    \centering
    \includegraphics[width=0.85\textwidth]{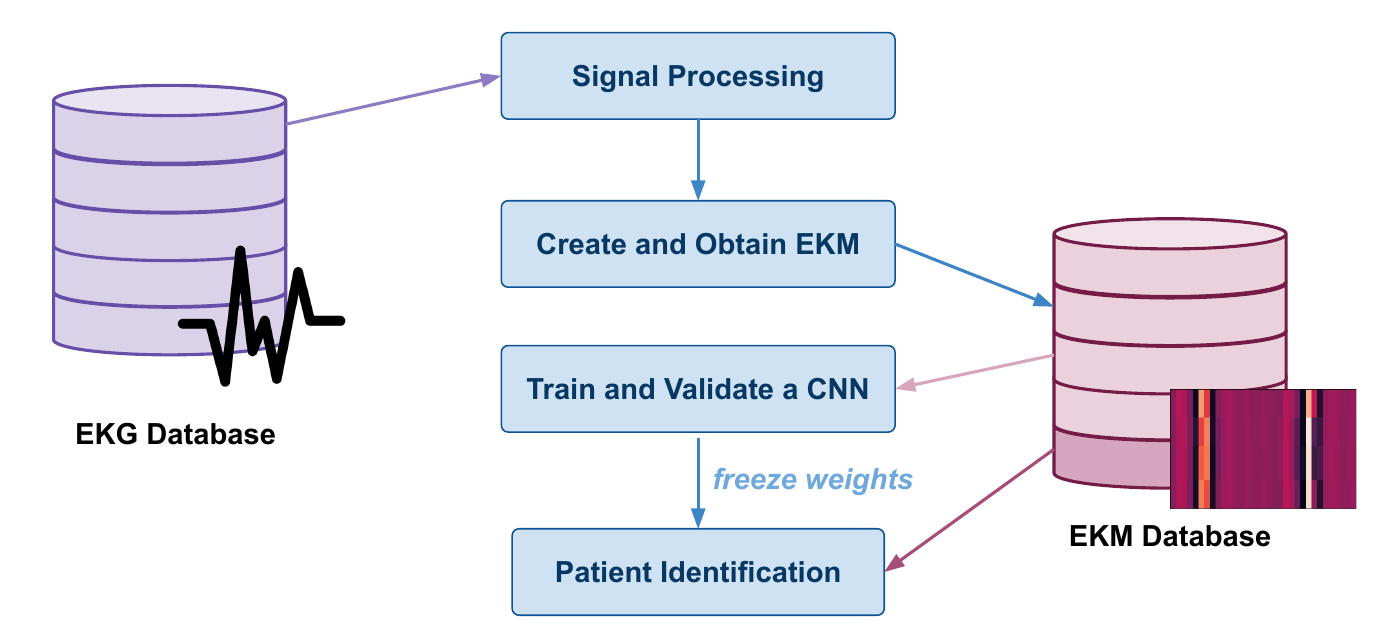}
    \caption{Pipeline of the methodology followed in \citep{ELEKTRA2022} from EKG {processing (noise reduction, peak detection, etc) to the creation of the EKM Database, training and validation of a CNN and final Patient Identification.}}
    \label{fig:pipeline}
\end{figure*}

The conversion of an \gls{ekg} into a heatmap involves several steps, as outlined in Figure \ref{fig:pipeline}, further described in~\citep{ELEKTRA2022} and available on Github at \url{https://github.com/cfusterbarcelo/ELEKTRA-approach}.

The process of creating the EKM dataset is presented in  Algorithm \ref{alg:EKM}. The initial step in performing patient identification or classification with EKGs is the obtaining of EKG recordings from the selected databases. This is achieved by reading the \texttt{.hea} and \texttt{.dat} files in which the \gls{ecg} is contained.

\begin{figure*}
    \begin{subfigure}{.3\textwidth}
        \centering
        \includegraphics[width=0.9\linewidth]{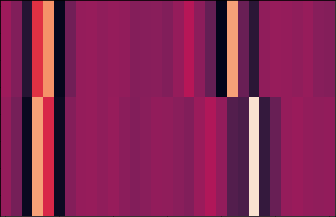}
        \caption{EKM to fed into the CNN with 3bpf}
        \label{sfig:2_ekm_3bpf}
    \end{subfigure}
    \hfill
    \begin{subfigure}{.3\textwidth}
        \centering
        \includegraphics[width=0.9\linewidth]{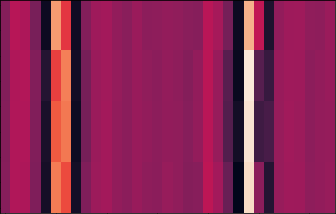}
        \caption{EKM to fed into the CNN with 5bpf}
        \label{sfig:2_ekm_5bpf}
    \end{subfigure}
    \hfill
    \begin{subfigure}{.3\textwidth}
        \centering
        \includegraphics[width=0.9\linewidth]{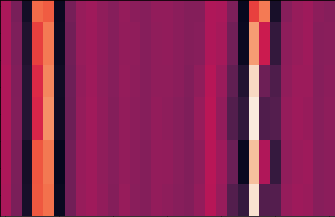}
        \caption{EKM to fed into the CNN with 7bpf}
        \label{sfig:2_ekm_7bpf}
    \end{subfigure}
    \caption{EKMs fed into the CNN depending on the bpf.}
    \label{fig:2_ekms}
\end{figure*}

Once the EKG has been read, certain parameters must be initialized. The sampling frequency (sf), {which defines how many samples per second are captured from teh ECG,} is specific to each database and is provided by the database. For example, the sf for the \gls{nsrdb} is 128 and for the \gls{mitdb} it is 360. The initial window {is a time segment of the ECG that will be processed, and it is updated as each EKM is generated. The initial window} is also initialized at this stage and will be updated after each EKM is obtained. This is necessary in order to move the beginning of the window by the number of \gls{bpf} that have been chosen as a hyperparameter ($bpf=(3,5,7)$). {The bpf is a key parameter that determines how many heartbeats are considered when creating each EKM image.} Resulting images depending on the \gls{bpf} are shown in Figure \ref{fig:2_ekms}.

\begin{algorithm}
\caption{Creation of an EKM database}
\label{alg:EKM}
\begin{algorithmic}
\STATE \textbf{Define path} to .hea and .dat files\;
\STATE \textbf{Initialise}: $Initial\, window = 0$, $sf$\;
\STATE \textbf{Define} Hyperparameters: $bpf$, $\alpha_i$, $\alpha_e$, train percentage, \#EKM total\;
\STATE R peaks list, filtered EKG $\leftarrow$ \textbf{PanTompkins}(unfiltered EKG, $sf$)\;
\STATE detrend EKG $\leftarrow$ \textbf{Detrend}(filtered EKG)\;
\STATE norm EKG $\leftarrow$ \textbf{Normalise}(detrend EKG)\;
\STATE $\mu$ = mean peak distance(R peaks list)\;
\WHILE{(train and test EKM < \#EKM total)}
    \STATE \textbf{\textit{Create EKM matrix}}($\mu$, R peaks list, norm EKG, Initial window, $\alpha_i$, $\alpha_e$)\;
    \STATE \textbf{Standarise} EKM$(-1,1)$\;
    \STATE \textbf{Generate} EKM image\;
    \STATE \textbf{Save} EKM in train or test\;
    \STATE $Initial\, window\, += bpf$\;
\ENDWHILE
\end{algorithmic}
\end{algorithm}

The values for $\alpha_i$ and $\alpha_e$ are the percentage of the signal that is to be taken before and after each peak of a window to construct each segment of the EKM. These values {control how much of the ECG signal around the R-peaks (the highest points in a heartbeat cycle) is used to form the segments for each EKM. These values} can be defined by the user, given certain thresholds ($\alpha_i < 100\% > \alpha_e$)\footnote{For this approach, $\alpha_i= 0.2$ and $\alpha_e= 1.3$, {ensuring that both the region before and after each R-peak is adequately captured, representing the full cardiac cycle}}. The total number of EKMs must be specified when dealing with an extensive database, such as the \gls{nsrdb}, as well as the percentage of the subsets for training and testing, are also determined.

Two lists are necessary for constructing each EKM: i) a clean and filtered EKG signal and ii) a list of the R peaks of this EKG signal. {The first list is the cleaned ECG signal after filtering out noise, and the second is a list of the R-peaks, which are used as reference points for aligning heartbeats.} These lists are obtained through the Pan and Tompkins algorithm~\citep{pantompkins}, {which is a widely used method to detect R-peaks in ECG signals}. Thus, the unfiltered EKG signal and the sampling frequency are provided to this algorithm, with the signal then being detrended. {Detrending removes any baseline drift from the ECG, making the signal more suitable for processing.} The mean distance between R peaks for each signal must also be calculated by using the list of R peaks obtained earlier.

With this information, the creation of all \gls{ekm} from the chosen database can proceed. { The process for creating each EKM involves using the cleaned ECG segments and aligning the R-peaks across a set of beats to form an image that can be fed into the CNN for classification.} The process for creating a single \gls{ekm} is detailed in Algorithm \ref{alg:oneEKM}. This involves creating a segment, based on the values of $\alpha$ and $\mu$, for each peak within the window, concatenating and aligning these segments, and repeating this process for the entire window.

\begin{algorithm}
\caption{Creation of \textbf{one} EKM}
\label{alg:oneEKM}
\begin{algorithmic}
\REQUIRE $\mu$, R peaks list, norm EKG, Initial window, $\alpha_i$, $\alpha_e$\;
\FOR{each peak $(p_x)$ of the current window}
    \STATE segment = norm EKG[$p_x - \alpha_i\mu:p_x + \alpha_e\mu$]\;
    \STATE All segments $\leftarrow$ Append(segment)\;
    \STATE EKM = Concatenate(All segments, 1)\;
\ENDFOR
\RETURN EKM\;
\end{algorithmic}
\end{algorithm}

Proceeding to Algorithm \ref{alg:EKM} following the creation of the \gls{ekm}, standardization of the \gls{ekm} is undertaken. {Standardization scales the values in the EKM to a consistent range (e.g., -1 to 1) so that they can be effectively processed by the CNN.} Subsequently, an image of the \gls{ekm} is generated by plotting it in the form of a heatmap and being saved. The process is repeated for subsequent windows until all the signal or the desired number of \gls{ekm}s are obtained.
The process of creating \gls{ekm}s is repeated for each user in the databases discussed in Section \ref{subsec:data}. For example, for the \gls{nsrdb} dataset, a specific number of images per user, such as 3000, are obtained by repeating the process. However, for the \gls{gudb}, \gls{ptbdb}, and \gls{mitdb} datasets, since the \gls{ekg}s for each user are shorter, the entire signal is used to extract as many \gls{ekm}s as possible. Therefore, the number of \gls{ekm}s may vary among datasets, users, and \gls{bpf}s.

Lastly, the division of each constructed dataset into training and testing sets is performed. Based on the number of images obtained from each database, a division of 80\% for training and 20\% for testing (as is the case for the \gls{nsrdb} ) or 90\% for training and 10\% for testing (as is the case for the other datasets) is established. Additionally, the training set is divided into a training and validation set, as is common practice, in order to cross-validate all parameters of the network. {To prevent any potential bias, the dataset was split based on users, meaning that all ECG images from a single user were exclusively used either in the training set or in the testing set, but never in both.}

\subsubsection{Convolutional Neural Network for patient identification}

\begin{figure}
    \centering
    \includegraphics[width=0.9\linewidth]{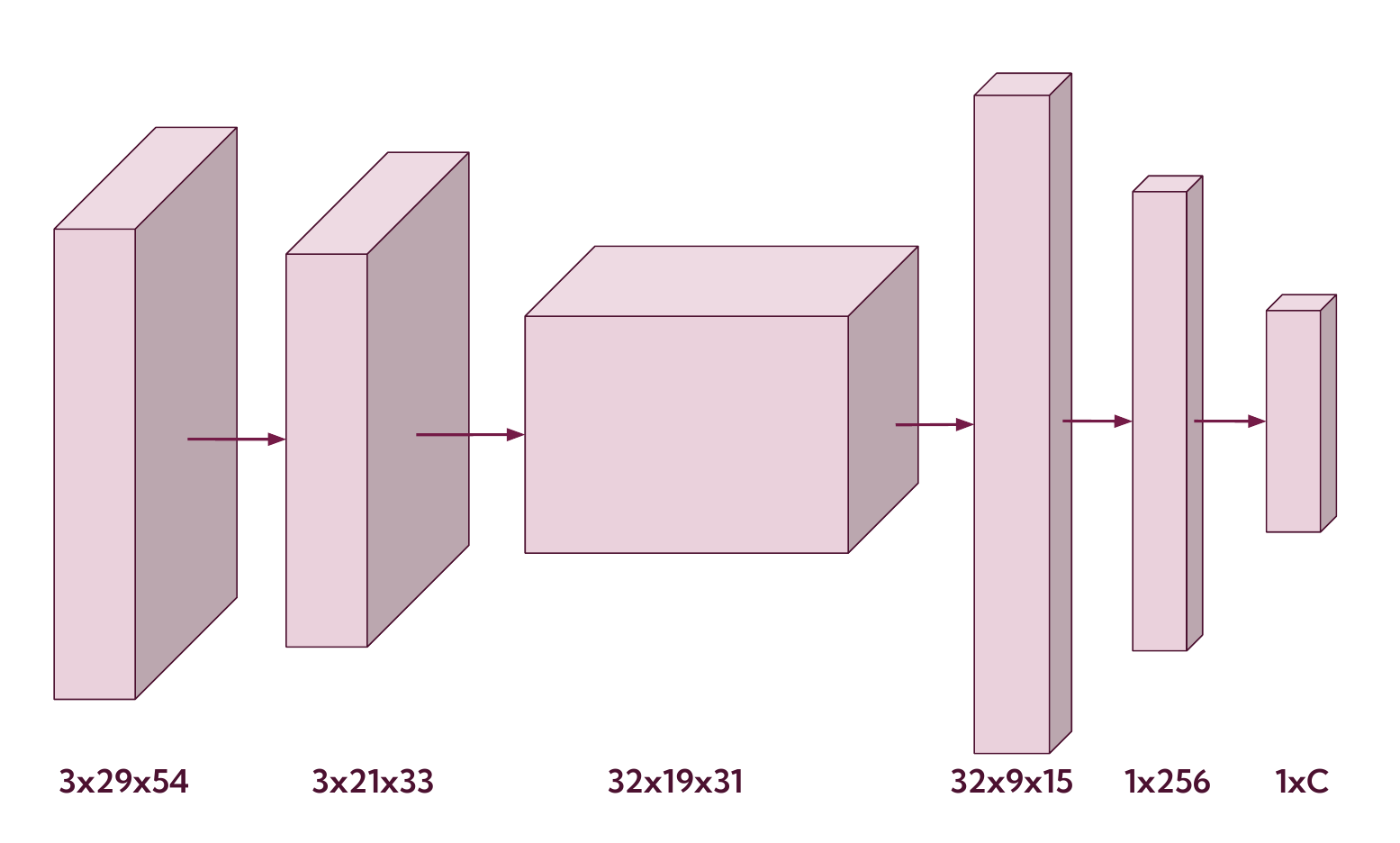}
    \caption{CNN Architecture. {\textit{C being the number of categories or classes.}}}
    \label{fig:cnn}
\end{figure}

The identification process is then carried out using the obtained \gls{ekm} datasets. The architecture of the proposed \gls{cnn} is detailed in Figure \ref{fig:cnn} and the layers of the networks with the number of parameters are detailed in Table \ref{tab:mod_sum}. A preprocessing step is performed as the first step by the network, where input images are cropped to reduce the number of parameters and eliminate irrelevant information. The cropped images are fed into a Convolutional Neural Network (CNN), and the second layer of the CNN is structured with a 2D Convolution, Rectified Linear Unit (Relu) activation, Max Pooling, and Dropout operations. The third layer, the fully connected layer (FCL), is the classification layer, which aims to group the features into the number of classes to be identified. The FCL consists of Flatten, Dense, and Softmax activation operations. The Adam optimizer is used for optimization, and the network is trained in batches with different numbers of epochs and steps per epoch based on the chosen database, with some experiments requiring up to 300 epochs.

\begin{table}[]
    \centering
    \caption{Model Summary of the CNN}
    \label{tab:mod_sum}
    \begin{tabular}{c c c}
    \toprule
        Layer (type) & Output Shape & Param \# \\
    \midrule
        Cropping2D & (None, 21, 33, 3) & 0 \\
        Conv2D & (None, 19, 31, 32) & 896\\
        MaxPooling2D & (None, 9, 15, 32) & 0 \\
        Dropout & (None, 9, 15, 32) & 0 \\
        Flatten & (None, 4320) & 0 \\
        Dense  & (None, 256) & 1,106,176\\
        Dense & (None, \textit{C}) & 59,624\\
    \bottomrule
    $^{\dagger}$Notation: \textit{C} number of users. 
    \end{tabular}

\end{table}

The \gls{cnn} trains the model by optimizing a Categorical Cross-Entropy cost function:
\begin{equation}
    H(y, \hat{y}) = -\sum_{c=1}^M y_{o,c} \log(p_{o,c})
\end{equation}
{where $M$ is the number of classes (i.e., the number of patients being identified), $y_{o,c}$ is a binary indicator (1 if class label $c$ is the correct classification for observation $o$, 0 otherwise), and $p_{o,c}$ is the predicted probability that observation $o$ belongs to class $c$.}

The number of epochs for the training process may vary for each experiment, depending on the model's response to the training. The batch size used for all experiments may also be subject to change depending on the size of the dataset. {After testing various configurations, a dropout rate of 0.7 was selected to prevent overfitting while maintaining performance. The learning rate was set to 0.003, which was determined through experimentation for optimal convergence. The Adam optimizer was chosen due to its adaptive learning rate and proven effectiveness in training convolutional neural networks.}

It is noteworthy that the primary goal of this study is to refine, enhance, execute, and assess the methodology presented in~\cite{ELEKTRA2022} for practical application, specifically in the area of patient identification. Using the databases incorporated in the experiments, it is demonstrated that patient identification can be achieved among a diverse range of patients and situations using a straightforward yet efficacious \gls{cnn}. The study highlights the ability to use a basic architecture to classify and identify patients with exceptional performance. The ultimate objective is to investigate the impact of various conditions and scenarios on patient identification and demonstrate the proposed approach's robustness.

\section{Results}
\subsection{Healthy Subjects}\label{subsec:nsrdb}
In this study, the database known as \gls{nsrdb} was used, as it comprises healthy individuals, as previously specified in Section \ref{subsec:data}. This database was chosen as a baseline due to two factors: first, it is the largest database in the possession of the researchers, which allows the extraction of 3000 images per user, resulting in a total of 54000 images per experiment regardless of the number of beats per frame (\gls{bpf}). Secondly, using healthy participants in this database provides an initial understanding of the model's performance in an ideal scenario.

As can be observed in Table \ref{tab:nsrdb-results}, the results of testing the model trained with \gls{nsrdb} are presented for different epochs. To demonstrate its performance, the model has been trained and evaluated for 3, 5, and 7 \gls{bpf}. It is noted that, when examining the results by epochs for each of the \gls{bpf} experiments, there is a trend in which the 150 epochs are the optimal choice. This is due to the fact that training the model with 100 epochs tends to result in underfitting, whereas training with 200 epochs results in overfitting. For instance, when considering the experiment's results with 5 \gls{bpf}, 99.78\% accuracy is obtained with 100 epochs. Then, with slightly more training, an even better result of 99.82\% accuracy is achieved with 150 epochs. However, further increasing the number of epochs to 200 results in a slight decrease in accuracy to 99.69\%, indicating that the network cannot continue learning features and suffers from overfitting.

It can be observed that the highest result is obtained using 7\gls{bpf} and 150 epochs, resulting in an accuracy of 99.84\%. Furthermore, it is noteworthy that all values for \gls{far} and \gls{frr} are low, indicating a high level of performance for the proposed method utilising the \gls{elektra} approach. It can be noted that the results obtained from this database are highly satisfactory, with near-perfect accuracy (approximately 100\%) and low error rates (approximately 0\%).  In conclusion, it can be stated that the presented method is a viable model capable of achieving low error rates and high accuracy in identifying healthy users, as demonstrated by using the \gls{nsrdb} database.

\begin{table}[]
\centering
\caption{ Results obtained from the experiments performed for the \textbf{\gls{nsrdb}}.}
\label{tab:nsrdb-results}
\begin{tabular}{ c c c c c c }
ine
\textbf{bpf}                 & \textbf{Epochs} & \textbf{Loss}   & \begin{tabular}[c]{@{}c@{}}\textbf{Accuracy}\\ (\%)\end{tabular} & \textbf{FAR}  (\%) & \textbf{FRR} (\%) \\\hline
                    & 100    & 0.0205 & 99.50          & 0.03      & 0.50     \\ 
                    & 150    & 0.0307 & 99.04         & 0.06      & 0.96     \\ 
\multirow{-3}{*}{3} & 200    & \textbf{0.0115} & \textbf{99.69}         & \textbf{0.02}     & \textbf{0.31}     \\\hline
                    & 100    & 0.0104 & 99.78         & 0.01      & 0.22     \\ 
                    & 150    & \textbf{0.0072} & \textbf{99.82}         & \textbf{0.01}      & \textbf{0.18}     \\ 
\multirow{-3}{*}{5} & 200    & 0.0123 & 99.69         & 0.02      & 0.31     \\\hline
                    & 100    & 0.0138 & 99.67         & 0.02      & 0.33     \\ 
                    & 150    & \textbf{0.0057} & \textbf{99.84}         & \textbf{0.01}      & \textbf{0.15}     \\ 
\multirow{-3}{*}{7} & 200    & 0.0078 & 99.84         & 0.01      & 0.16     \\\hline
\end{tabular}
\end{table}

\subsection{Patients with Arrhythmia and healthy subjects}\label{subsec:mitdb}

The \gls{mitdb} has been selected for analysis to study the proposed method's behaviour when applied to a diverse population of individuals, including those with and without significant arrhythmia. Hence, one \gls{cvd} is analysed together with healthy subjects in this experiment. This experiment aims to determine whether cardiovascular disease, such as arrhythmia, can negatively impact the identification results when both affected and healthy individuals are included. This examination provides a closer approximation to the real-world scenario in a healthcare facility.

It is noted that the number of images or \gls{ekm}s available for the \gls{mitdb} is less than that for \gls{nsrdb}. As the number of \gls{bpf} increases, the number of images decreases as shown in Table \ref{tab:mitdb}. For example, the number of images for 3\gls{bpf} is 35949, while for 5\gls{bpf} and 7\gls{bpf} it is 21149 and 15119 respectively, resulting in approximately half the number of images for 7\gls{bpf} compared to 3\gls{bpf}.

It should be noted that despite the decrease in the number of images compared to the \gls{nsrdb}, promising results have been obtained when utilising the \gls{mitdb}. As previously mentioned, it has been observed that there is a slight overfitting of the network occurs when training with 200 epochs for some experiments utilising different numbers of \gls{bpf}. However, it is also observed that for the 7 \gls{bpf} case, 200 epochs of training yields the best result for the entire database. Generally, the best results are obtained when using 5 \gls{bpf}, with the highest accuracy of 97.89\% being achieved when using 7 \gls{bpf} and 200 epochs of training.

It can be inferred from the results obtained from the \gls{mitdb}, where healthy users are mixed with patients with various cardiovascular diseases, that the proposed approach can satisfactorily classify and identify patients regardless of their cardiac health condition. The accuracy obtained, which exceeds 95\% and reaches 97.89\%, suggests that the method is highly effective. Furthermore, the low error rates represented by the \gls{far} and \gls{frr} values indicate that the proposed method can be successfully employed for patient identification healthcare facilities regardless of the cardiac health condition of each patient.

\begin{table}[]
\centering
\caption{Results obtained from the experiments conducted on \textbf{\gls{mitdb}}.}
\label{tab:mitdb}
\resizebox{1\columnwidth}{!}{
\begin{tabular}{ c c c c c c c }
\hline
\multicolumn{1}{ c }{\textbf{bpf}} & \multicolumn{1}{c }{\textbf{EKMs}} & \textbf{Epochs} & \textbf{Loss}   & \begin{tabular}[c]{@{}c@{}}\textbf{Accuracy}\\ (\%)\end{tabular} & \begin{tabular}[c]{@{}c@{}}\textbf{FAR}\\ (\%)\end{tabular} & \begin{tabular}[c]{@{}c@{}}\textbf{FRR}\\ (\%)\end{tabular} \\\hline
\multicolumn{1}{ c }{}                            & \multicolumn{1}{c }{}                             & 150    & 0.1268 & 96.88                                                   & 0.07                                               & 3.12                                               \\ 
\multicolumn{1}{ c }{}                            & \multicolumn{1}{c }{}                             & 200    & \textbf{0.1160} & \textbf{96.80}                                                   & \textbf{ 0.07 }                                               & \textbf{3.20}                                               \\  
\multicolumn{1}{ c }{\multirow{-3}{*}{3}}         & \multicolumn{1}{c }{\multirow{-3}{*}{35949}}      & 250    & 0.1136 & 96.38                                                   & 0.71                                               & 3.17                                               \\\hline
                                                  &                                                   & 150    & 0.1762 & 96.12                                                   & 0.08                                               & 3.88                                               \\  
                                                  &                                                   & 200    & 0.1071 & 97.63                                                   & 0.05                                               & 2.37                                               \\ 
\multirow{-3}{*}{5}                               & \multirow{-3}{*}{21149}                           & 250    & \textbf{0.1004} & \textbf{97.89 }                                                  & \textbf{0.07 }                                              & \textbf{3.27}                                               \\\hline
                                                  &                                                   & 150    & 0.1551 & 95.92                                                   & 0.09                                               & 4.08                                               \\  
                                                  &                                                   & 200    & 0.1277 & 96.12                                                   & 0.08                                               & 3.88                                               \\  
\multirow{-3}{*}{7}                               & \multirow{-3}{*}{15119}                           & 250    & \textbf{0.0934} & \textbf{97.89}                                                  & \textbf{0.04}                                               & \textbf{2.11}                                               \\\hline
\end{tabular}
}
\end{table}

\subsection{Patients with various CVD and healthy subjects}\label{subsec:ptbdb}
An important characteristic of the \gls{ptbdb} is highlighted in Section \ref{subsec:data} as it contains both individuals with and without \gls{cvd}. This database has been constructed in such a way that it allows the exclusion of certain users based on the specific \gls{cvd} they possess. This feature enables the examination of the performance of the proposed patient identification methodology on a dataset comprising solely of individuals with \gls{cvd} and comparison with other databases with diverse subjects to continue with the presented application for a healthcare facility.

Two separate experiments were conducted on this database: one including all users and another one including only those with certain types of cardiac disease and excluding healthy subjects. The results of these experiments can be observed in Table \ref{tab:ptbdb-results}. The first experiment includes a variety of cardiac diseases, such as Myocarditis and Dysrhythmia, and healthy subjects, providing a mixed population similar to the experiments conducted on the \gls{mitdb}. However, due to the limited size of the \gls{ekg} recordings in this database, the number of \gls{ekm}s obtained is smaller. Although two hundred thirty-two users are comprised in this database. 

It can be inferred from the results obtained in the experiment conducted on the \gls{ptbdb} that, despite the drastic reduction in the number of \gls{ekm}s and the increase in the number of users (4.8 times more users than the \gls{mitdb}), the proposed method is capable of identifying users with high performance. The accuracy achieved, up to 93.96\% for 3\gls{bpf}, is noteworthy considering the limitations in the number of images and the increase in the number of users. This demonstrates the feasibility and robustness of the proposed method.

Based on the results obtained from the entire processing of \gls{ptbdb}, it can be concluded that the proposed method \gls{elektra} is capable of effectively identifying patients, even in situations where a wide range of cardiovascular disorders are present among the user population. Despite using a database with a smaller number of images and an increased number of users, the method demonstrates promising performance in terms of accuracy. Furthermore, while the \gls{far} values are low, indicating a high performance in correctly identifying legitimate users, the \gls{frr} values are not as favourable, indicating that the method may sometimes reject legitimate users. However, considering the primary objective of patient identification in a hospital setting and the need to prevent impersonation, the results can still be considered satisfactory.

\begin{table}[]
\centering
\caption{Results obtained during the experiments carried out with \textbf{\gls{ptbdb}} with 3, 5 and 7\gls{bpf}. }
\label{tab:ptbdb-results}
\resizebox{1\columnwidth}{!}{
\begin{tabular}{ c c c c c c c }
ine
\textbf{bpf} &
  \textbf{EKMs} &
  \textbf{Epochs} &
  \textbf{Loss} &
  \begin{tabular}[c]{@{}c@{}}\textbf{Accuracy} \\ (\%)\end{tabular} &
  \begin{tabular}[c]{@{}c@{}}\textbf{FAR}\\  (\%)\end{tabular} &
  \begin{tabular}[c]{@{}c@{}}\textbf{FRR}\\ (\%)\end{tabular} \\\hline
                    &                        & 150 & 0.4454 & 88.55 & 0.05 & 11.45  \\   
                    &                        & 200 & 0.4214 & 89.80  & 0.05 & 10.20  \\   
\multirow{-3}{*}{3} & \multirow{-3}{*}{9854} & 250 & \textbf{0.2831} & \textbf{93.96} & \textbf{0.03} & \textbf{6.04}  \\\hline
                    &                        & 150 & 0.7463 & 82.85 & 0.08 & 17.15  \\   
                    &                        & 200 & 0.512  & 87.27 & 0.06 & 12.73  \\   
\multirow{-3}{*}{5} & \multirow{-3}{*}{5891} & 250 & \textbf{0.4594} & \textbf{87.44} & \textbf{0.06} & \textbf{12.56} \\\hline
                    &                        & 150 & 1.2036 & 71.98 & 0.15 & 28.02  \\   
                    &                        & 200 & 0.9239 & 79.12 & 0.11 & 20.88  \\   
\multirow{-3}{*}{7} & \multirow{-3}{*}{4180} & 250 & \textbf{0.5128} & \textbf{87.64} & \textbf{0.07} & \textbf{12.36} \\\hline
\end{tabular}%
}
\end{table}

The results of the experiments carried out on the segmented \gls{ptbdb} database, which consists only of users with \gls{cvd}, are presented in Table \ref{tab:ptbdb-cvd-results}. It is observed that better results are obtained with this segmentation of the database than with the whole database. This may be due to the increased dissimilarity of the \gls{ecg} recordings between users and \gls{cvd}, as previously demonstrated in other studies(\cite{MONEDERO2022104536, UBEYLI20081196, LEE2018S121, ecm_2022, STROKEAHA, EBRAHIMI2020100033}), making identification easier in this segmented population. It is worth noting that with a relatively small database of 162 users, the proposed approach demonstrates its potential for accurately identifying patients with \gls{cvd}.

\begin{table}[]
\centering
\caption{Results obtained when testing over the \textbf{\gls{ptbdb}} for patients with \gls{cvd}.}
\label{tab:ptbdb-cvd-results}
\resizebox{1\columnwidth}{!}{
\begin{tabular}{ c c c c c c c }
ine
{ \textbf{bpf}} &
  { \textbf{EKMs}} &
  { \textbf{Epochs}} &
  { \textbf{Loss}} &
  { \begin{tabular}[c]{@{}c@{}}\textbf{Accuracy}\\ (\%)\end{tabular}} &
  { \begin{tabular}[c]{@{}c@{}}\textbf{FAR}\\ (\%)\end{tabular}} &
  { \begin{tabular}[c]{@{}c@{}}\textbf{FRR}\\ (\%)\end{tabular}} \\\hline
                    &                        & 150 & 0.4445 & 89.61 & 0.07 & 10.39 \\   
                    &                        & 200 & 0.2703 & 93.91 & 0.04 & 6.09  \\   
                    &                        & 250 & 0.2302 & 96.40  & 0.02 & 3.60   \\   
\multirow{-4}{*}{3} & \multirow{-4}{*}{7266} & 300 & \textbf{0.1920}  & \textbf{97.09} & \textbf{0.02} & \textbf{2.91}  \\\hline
                    &                        & 150 & 0.6338 & 84.13 & 0.10  & 15.87 \\   
                    &                        & 200 & 0.4225 & 88.91 & 0.01 & 11.09 \\   
                    &                        & 250 & 0.3125 & 93.04 & 0.05 & 6.96  \\   
\multirow{-4}{*}{5} & \multirow{-4}{*}{4350} & 300 & \textbf{0.2124} & \textbf{95.00 }   & \textbf{0.03} & \textbf{5.00}     \\\hline
                    &                        & 150 & 1.4065 & 65.10  & 0.25 & 34.9  \\   
                    &                        & 200 & 0.8560  & 81.21 & 0.13 & 18.79 \\   
                    &                        & 250 & \textbf{0.5645 }& \textbf{88.26} & \textbf{0.09} & \textbf{11.74} \\   
\multirow{-4}{*}{7} & \multirow{-4}{*}{3066} & 300 & 0.5114 & 86.58 & 0.10  & 13.42 \\\hline
\end{tabular}%
}
\end{table}

Based on the experiments conducted on the \gls{ptbdb}, it can be concluded that the proposed approach for patient identification can accurately identify a wide range of individuals with and without various cardiac conditions. Furthermore, the results obtained when only individuals with cardiac conditions were considered to demonstrate the model's enhanced ability to identify this population, potentially due to the inter-subject variability present among individuals with cardiac conditions. These findings showcase the feasibility and robustness of the proposed model when applied to real-world scenarios in healthcare facilities.

\subsection{Subjects performing activities}\label{subsec:gudb}

In this experiment, the chosen database was used to understand the feasibility and robustness of the proposed approach for patient identification in different scenarios where cardiovascular activity can affect the identification process. In a practical scenario, it may occur that patients exhibit varying heartbeat rates, which does not necessarily impact their identification. For instance, a patient who has recently been involved in an accident or who is walking through a hospital corridor is expected to exhibit a higher heart rate than when they are resting in bed. As such, a study of the impact of cardiovascular activity and differing heartbeat rates on patient identification is deemed necessary. For this purpose, the \gls{gudb} is studied, as it encompasses users who are engaging in different activities with varying heartbeat rates.

It is worth noting that the number of \gls{ekm}s obtained to perform patient identification in this database is relatively low, with 25 healthy users in five different scenarios as described in Section \ref{subsec:data}.  \gls{ecg} recordings of each participant were taken while they participated in five different activities. It is recognised that the heart rate and behaviour of the individual would differ in each of these activities. For example, sitting would be considered an activity with a low heart rate, whereas running would be considered a cardiovascular activity with a higher heart rate. Therefore, the main objective of using this database was to analyse the performance of the proposed approach for patient identification in scenarios where the individual's heart rate may vary as in a real application for healthcare systems and facilities.

The experiments carried out in this section have been grouped into three main categories: those in which all activities are separated and images containing 3 \gls{bpf} are used, those in which the same approach is taken but with 5 \gls{bpf}, and those in which all activities are combined to train and test the network to simulate better real-life scenarios where a patient may have varying heart rate during the identification process.

Based on the experimentation conducted on the \gls{gudb} with 3\gls{bpf}, as presented in Table \ref{tab:gudb-3bpf}, it can be inferred that the proposed approach for patient identification is capable of achieving a high level of accuracy, even with a low number of images. Furthermore, the results obtained per scenario suggest that the identification of users with lower heart rates may be more successful than those with higher heart rates. This observation supports the potential utility of the proposed approach in real-world scenarios where patients may have varying heart rates during identification.

\begin{table}[]
\centering
\caption{Results obtained from experiments carried out over \textbf{\gls{gudb}} with \textbf{3\gls{bpf}} and over the five different scenarios. }
\label{tab:gudb-3bpf}
\resizebox{1\columnwidth}{!}{
\begin{tabular}{ c c c c c c c }
ine
\textbf{Scenario} &
  \textbf{EKMs} &
  \textbf{Epochs} &
  \textbf{Loss} &
  \begin{tabular}[c]{@{}c@{}}\textbf{Accuracy}\\ (\%)\end{tabular} &
  \begin{tabular}[c]{@{}c@{}}\textbf{FAR}\\ (\%)\end{tabular} &
  \begin{tabular}[c]{@{}c@{}}\textbf{FRR}\\ (\%)\end{tabular} \\\hline
                          &                        & 150 & 0.2721 & 94.87 & 0.21 & 5.13  \\  
                          &                        & 200 & \textbf{0.2504} & \textbf{95.51} & \textbf{0.19} & \textbf{4.49}  \\ 
\multirow{-3}{*}{\begin{tabular}[c]{@{}c@{}}Hand-\\ bike\end{tabular}} &
  \multirow{-3}{*}{1529} &
  250 &
  0.1918 &
  94.23 &
  0.24 &
  5.77 \\\hline
                          &                        & 150 & 0.8128 & 74.58 & 1.06 & 25.42 \\   
                          &                        & 200 & \textbf{0.6302} & \textbf{82.63} & \textbf{0.72} & \textbf{17.37} \\   
\multirow{-3}{*}{Jogging} & \multirow{-3}{*}{2205} & 250 & 0.5329 & 82.63 & 0.72 & 17.37 \\\hline
                          &                        & 150 & 0.1358 & 94.31 & 0.24 & 5.69  \\   
                          &                        & 200 & \textbf{0.0798} & \textbf{99.19} & \textbf{0.03} & \textbf{0.81}  \\   
\multirow{-3}{*}{Sitting} & \multirow{-3}{*}{1335} & 250 & 0.0907 & 94.56 & 0.10  & 2.44  \\\hline
                          &                        & 150 & 0.2008 & 93.66 & 0.26 & 6.34  \\   
                          &                        & 200 & 0.0867 & 97.89 & 0.09 & 2.11  \\   
\multirow{-3}{*}{Walking} & \multirow{-3}{*}{1556} & 250 & \textbf{0.0568} & \textbf{98.59} & \textbf{0.06} & \textbf{1.41}  \\\hline
                          &                        & 150 & 0.2001 & 93.33 & 0.28 & 6.67  \\   
                          &                        & 200 & \textbf{0.1803} & \textbf{94.00}  & \textbf{0.25} & \textbf{6.00}     \\   
\multirow{-3}{*}{Maths}   & \multirow{-3}{*}{1474} & 250 & 0.1254 & 94.00  & 0.25 & 6.00     \\\hline
\end{tabular}%
}
\end{table}

Following the experiments performed on the \gls{gudb} with different numbers of \gls{bpf}, it can be observed that the \gls{bpf} is a critical parameter that must be carefully considered when dealing with different heart rates. The results obtained in tables \ref{tab:gudb-3bpf} and \ref{tab:gudb-5bpf} suggest that it may be easier to identify users at rest with lower heart rates than when their heart beats faster. This could also explain the results obtained for the other activities studied in the tables. However, it is important to note that the number of images may also play a role in the performance of the network, as less than a thousand images may not be sufficient to properly train and test the network. As a result, the experiments for the \gls{gudb} with 7 \gls{bpf} were not included due to the limited number of images available.

The results obtained in the experimentation of the \gls{gudb} when all activities are merged, as presented in Table \ref{tab:gudb-allvsall}, demonstrate the feasibility of the proposed approach in identifying patients with different heartbeat rates. The highest accuracy of 91.32\% was achieved in the experiments with 3\gls{bpf}, 250 epochs, and 8099 images. A comparison with the results obtained in the experiment where the scenarios were separated, as presented in Table \ref{tab:gudb-3bpf}, reveals that this new result is better than those obtained in scenarios involving cardiac activity (such as jogging with an 85.82\%), but not as high as resting scenarios (such as sitting with a 98.51\%).

The results obtained with 5\gls{bpf} show lower accuracy and error rates, which may be attributed to the reduced number of images used compared to the experiment with 3\gls{bpf}. This highlights the importance of enroling patients in different situations and cardiac conditions to improve the performance of the proposed approach in identifying patients regardless of their heartbeat rhythm at the time of identification.

The robustness of the presented approach for patient identification in different scenarios and with varying heart rates is demonstrated through experimentation on the \gls{gudb}. The results obtained in this experiment indicate the feasibility of identifying patients in different situations, such as those that may occur in a healthcare facility where patients may arrive with elevated heart rates due to accidents or other reasons. Besides, it is crucial to consider the possibility of alterations in the heart rhythm patterns of patients during their hospitalisation as a result of their daily activities. This experimentation brings the proposed approach closer to real-life applications in healthcare systems and facilities.

\begin{table}[]
\centering
\caption{Results obtained from experiments carried out on \textbf{\gls{gudb}} with \textbf{5\gls{bpf}} and in the five different scenarios. }
\label{tab:gudb-5bpf}
\resizebox{1\columnwidth}{!}{
\begin{tabular}{ c c c c c c c }
ine
\textbf{Scenario} &
  \textbf{EKMs} &
  \textbf{Epochs} &
  \textbf{Loss} &
  \begin{tabular}[c]{@{}c@{}}\textbf{Accuracy}\\ (\%)\end{tabular} &
  \begin{tabular}[c]{@{}c@{}}\textbf{FAR}\\ (\%)\end{tabular} &
  \begin{tabular}[c]{@{}c@{}}\textbf{FRR}\\ (\%)\end{tabular} \\\hline
                          &                        & 150 & 0.4945 & 83.33 & 0.69 & 16.67 \\   
                          &                        & 200 & \textbf{0.359}  & \textbf{90.74} & \textbf{0.39} & \textbf{9.26}  \\   
\multirow{-3}{*}{\begin{tabular}[c]{@{}c@{}}Hand-\\ bike\end{tabular}} &
  \multirow{-3}{*}{914} &
  250 &
  0.385 &
  87.96 &
  0.5 &
  12.04 \\\hline
                          &                        & 150 & 0.7453 & 79.10 & 0.91 & 20.9  \\   
                          &                        & 200 & 0.6248 & 83.58 & 0.71 & 16.42 \\   
\multirow{-3}{*}{Jogging} & \multirow{-3}{*}{1317} & 250 & \textbf{0.6068} & \textbf{85.82} & \textbf{0.62} & \textbf{14.18} \\\hline
                          &                        & 150 & 0.2813 & 95.52 & 0.19 & 4.48  \\   
                          &                        & 200 & 0.1683 & 97.01 & 0.13 & 2.99  \\   
\multirow{-3}{*}{Sitting} & \multirow{-3}{*}{792}  & 250 & \textbf{0.1221} & \textbf{98.51} & \textbf{0.06} & \textbf{1.49}  \\\hline
                          &                        & 150 & 0.3204 & 93.10 & 0.29 & 6.9   \\   
                          &                        & 200 & \textbf{0.2077} & \textbf{95.40}  & \textbf{0.19} & \textbf{4.60}   \\   
\multirow{-3}{*}{Walking} & \multirow{-3}{*}{912}  & 250 & 0.2821 & 90.80 & 0.38 & 9.20   \\\hline
                          &                        & 150 & 0.3690 & 90.32 & 0.40  & 9.68  \\   
                          &                        & 200 & 0.1891 & 94.62 & 0.22 & 5.38  \\   
\multirow{-3}{*}{Maths}   & \multirow{-3}{*}{881}  & 250 & \textbf{0.1316} & \textbf{97.85} & \textbf{0.09} & 2.15  \\\hline
\end{tabular}%
}
\end{table}

\begin{table}[]
\centering
\caption{Results obtained from the experiments performed over the \gls{gudb} with 3 and 5\gls{bpf} joining all scenarios by doing an All vs. All. }
\label{tab:gudb-allvsall}
\resizebox{1\columnwidth}{!}{
\begin{tabular}{ c c c c c c c }
ine
\textbf{bpf} &
  \textbf{EKMs} &
  \textbf{Epochs} &
  \textbf{Loss} &
  \begin{tabular}[c]{@{}c@{}}\textbf{Accuracy}\\ (\%)\end{tabular} &
  \begin{tabular}[c]{@{}c@{}}\textbf{FAR}\\ (\%)\end{tabular} &
  \begin{tabular}[c]{@{}c@{}}\textbf{FRR}\\ (\%)\end{tabular} \\\hline
                    &                        & 150 & 0.3444 & 89.06 & 0.45 & 11.62 \\   
                    &                        & 200 & 0.3441 & 88.81 & 0.47 & 11.49 \\   
\multirow{-3}{*}{3} & \multirow{-3}{*}{8099} & 250 & \textbf{0.2963} & \textbf{91.32} & \textbf{0.36} & \textbf{9.06}  \\\hline
                    &                        & 150 & 0.7209 & 79.19 & 0.87 & 20.81 \\   
                    &                        & 200 & 0.6205 & 81.89 & 0.75 & 18.11 \\   
\multirow{-3}{*}{5} & \multirow{-3}{*}{4816} & 250 & \textbf{0.5708} & \textbf{82.47} & \textbf{0.73} & \textbf{17.53} \\\hline
\end{tabular}%
}
\end{table}

\section{Discussions \& Conclusions}

{This paper presents a novel approach for patient identification in healthcare systems using electrocardiogram (ECG) signals. A convolutional neural network (CNN) is used to classify users based on electrocardiomatrices (EKMs), a specific type of image derived from ECG signals. The proposed identification system is evaluated in multiple databases, providing a comprehensive understanding of its potential in real-world scenarios. The study demonstrates the feasibility and robustness of this approach and its ability to overcome the limitations of current patient identification methods in healthcare\footnote{A detailed comparison between our proposal and ECG-based identification systems has been included in the Appendix section for thoroughness.}. The results of the experiments indicate that the proposed system has the potential to be a reliable and effective method for use in healthcare facilities, and it is inclusive even when patients have health conditions or impairments.}  
The results of the experiment carried out in this study provide evidence of the effectiveness and robustness of the proposed approach for the identification of patients using \gls{ecg}s. As outlined in~\cite{noran2022towards}, the presented method meets several essential requirements for a patient identification system. The experimentation conducted in Section \ref{subsec:gudb}, where the approach was tested on subjects performing different activities, demonstrates the potential applicability of the proposed method in various scenarios, including emergencies. Furthermore, the proposed solution is highly maintainable, as a fine-tuning of the network would be sufficient to adapt to new situations, cardiac diseases or activities. Additionally, the use of \gls{ecg}s as a signal for patient identification is already well-established in healthcare systems and facilities, potentially increasing acceptance among both patients and medical professionals, as the signal is already a commonly used diagnostic tool. This makes the approach easy to learn and operate, making it a valuable addition to healthcare systems and facilities.
The reliability and effectiveness of the proposed approach for patient identification using \gls{ecg} signals has been demonstrated through experimentation conducted on both individuals with and without cardiovascular conditions, as evidenced in the examination of the \gls{mitdb} and \gls{ptbdb} datasets (presented in Sections \ref{subsec:mitdb} and \ref{subsec:ptbdb}). The use of this methodology for the study and analysis of various cardiovascular diseases has been established through previous works (such as those presented in~\cite{ecm, ecm_2022, salinas2021detection}), thereby indicating the possibility of both diagnosing cardiovascular conditions and identifying patients through the use of this approach. Other studies over user or patient identification with \gls{ecg} signals do not test or study their methodology over users with different \gls{cvd}. Thus, including people with cardiovascular diseases in the identification process has been shown to be possible in the presented research, making the method inclusive for a diverse range of patients as everyone has a beating heart and \gls{cvd} does not affect the patient identification. Additionally, the proposed approach does not require the individual's conscious participation, making it suitable for identifying unconscious patients or those with varying heart rates or cardiovascular conditions. Finally, it is noted that biometric systems utilising electrocardiogram (ECG) signals for patient identification possess a significant advantage over other biometric characteristics, as the diagnosis of the patient's cardiovascular health can also be obtained during the identification process.

{While the proposed approach has shown promising results, there are some limitations that must be acknowledged. First, the study was conducted using publicly available databases, which, although diverse, may not fully represent all real-world clinical scenarios. Additionally, while the model demonstrated high performance across different datasets, its robustness in identifying patients with rare cardiovascular conditions or severe signal artifacts needs further exploration. Another limitation is the lack of long-term studies that assess the system's performance over extended periods, such as several months or years. This is particularly important for elderly individuals whose ECG may vary due to disease progression, and for athletes whose ECG may fluctuate significantly between training and competition phases. Due to the scarcity of suitable longitudinal datasets, this aspect could not be addressed in the current work. Future research should focus on collecting and evaluating long-term ECG data to better understand these dynamics. Furthermore, while the model performs well in scenarios with varying heart rates, additional optimization might be needed to handle extreme physiological conditions, such as arrhythmias during high physical exertion.}

{In future work, we aim to focus on the practical implementation of this system in real-world healthcare environments. This will involve integrating the proposed model into a standalone machine or device that can be deployed in hospitals and clinics for automatic patient identification. Key aspects of this development will include optimizing the system for real-time performance, ensuring compatibility with existing healthcare infrastructure, and conducting further testing in live clinical settings to validate the model's effectiveness under diverse conditions.}

In summary, the results of the experiments carried out in this study indicate that the proposed approach to the identification of patients using \gls{ecg}s has the potential to be a reliable and effective method for use in healthcare facilities and systems. The use of \gls{ecg}s as a signal for patient identification is well-established. In this sense, the proposed approach based on ECG signal is suitable for different scenarios and possesses a high degree of maintainability. These findings indicate that further research and development of this approach may be merited to bring it closer to real-world application in healthcare systems and facilities.

\bibliographystyle{IEEEtran}
\bibliography{bibliography}

\begin{thebibliography}{10}
\providecommand{\url}[1]{#1}
\csname url@samestyle\endcsname
\providecommand{\newblock}{\relax}
\providecommand{\bibinfo}[2]{#2}
\providecommand{\BIBentrySTDinterwordspacing}{\spaceskip=0pt\relax}
\providecommand{\BIBentryALTinterwordstretchfactor}{4}
\providecommand{\BIBentryALTinterwordspacing}{\spaceskip=\fontdimen2\font plus
\BIBentryALTinterwordstretchfactor\fontdimen3\font minus \fontdimen4\font\relax}
\providecommand{\BIBforeignlanguage}[2]{{%
\expandafter\ifx\csname l@#1\endcsname\relax
\typeout{** WARNING: IEEEtran.bst: No hyphenation pattern has been}%
\typeout{** loaded for the language `#1'. Using the pattern for}%
\typeout{** the default language instead.}%
\else
\language=\csname l@#1\endcsname
\fi
#2}}
\providecommand{\BIBdecl}{\relax}
\BIBdecl

\bibitem{alkhaqani2023patient}
A.~L. Alkhaqani, ``Patient identification errors in the hospital setting: A prospective observational study,'' \emph{Al-Rafidain Journal of Medical Sciences (ISSN: 2789-3219)}, vol.~4, pp. 1--5, 2023.

\bibitem{nigam2022biometric}
D.~Nigam, S.~N. Patel, P.~Raj~Vincent, K.~Srinivasan, and S.~Arunmozhi, ``Biometric authentication for intelligent and privacy-preserving healthcare systems,'' \emph{Journal of Healthcare Engineering}, vol. 2022, 2022.

\bibitem{aguilar2006positive}
A.~Aguilar, W.~Van Der~Putten, and F.~Kirrane, ``Positive patient identification using rfid and wireless networks,'' in \emph{HISI 11th Annual Conference and Scientific Symposium}.\hskip 1em plus 0.5em minus 0.4em\relax Citeseer, 2006.

\bibitem{mccoy2013matching}
A.~B. McCoy, A.~Wright, M.~G. Kahn, J.~S. Shapiro, E.~V. Bernstam, and D.~F. Sittig, ``Matching identifiers in electronic health records: implications for duplicate records and patient safety,'' \emph{BMJ quality \& safety}, vol.~22, no.~3, pp. 219--224, 2013.

\bibitem{riplinger2020patient}
L.~Riplinger, J.~Piera-Jim{\'e}nez, and J.~P. Dooling, ``Patient identification techniques--approaches, implications, and findings,'' \emph{Yearbook of medical informatics}, vol.~29, no.~01, pp. 081--086, 2020.

\bibitem{patientid_executivesummary}
E.~I. P.~D. Dive, ``Patient identification: Executive summary,'' 2016.

\bibitem{patientnow}
\BIBentryALTinterwordspacing
A.~Krzepicki, ``New perspectives on the patient id problem in healthcare,'' \emph{Patient Now}, November 2022. [Online]. Available: \url{http://patientidnow.org/wp-content/uploads/2022/11/PIDN-Research-Findings-Final.pdf}
\BIBentrySTDinterwordspacing

\bibitem{lippi2017managing}
G.~Lippi, C.~Mattiuzzi, C.~Bovo, and E.~J. Favaloro, ``Managing the patient identification crisis in healthcare and laboratory medicine,'' \emph{Clinical biochemistry}, vol.~50, no. 10-11, pp. 562--567, 2017.

\bibitem{hathaliya2020securing}
J.~J. Hathaliya, S.~Tanwar, and R.~Evans, ``Securing electronic healthcare records: A mobile-based biometric authentication approach,'' \emph{Journal of Information Security and Applications}, vol.~53, p. 102528, 2020.

\bibitem{hemesath2015educational}
M.~P. Hemesath, H.~B.~d. Santos, E.~M.~S. Torelly, A.~d.~S. Barbosa, and A.~M. M.~d. Magalh{\~a}es, ``Educational strategies to improve adherence to patient identification,'' \emph{Revista Ga{\'u}cha de Enfermagem}, vol.~36, pp. 43--48, 2015.

\bibitem{tase2015patient}
T.~H. Tase and D.~M.~R. Tronchin, ``Patient identification systems in obstetric units, and wristband conformity,'' \emph{Acta Paulista de Enfermagem}, vol.~28, pp. 374--380, 2015.

\bibitem{renner1993wristband}
S.~Renner, P.~Howanitz, and P.~Bachner, ``Wristband identification error reporting in 712 hospitals. a college of american pathologists' q-probes study of quality issues in transfusion practice.'' \emph{Archives of pathology \& laboratory medicine}, vol. 117, no.~6, pp. 573--577, 1993.

\bibitem{de2019interventions}
H.~A. De~Rezende, M.~M. Melleiro, and G.~T. Shimoda, ``Interventions to reduce patient identification errors in the hospital setting: a systematic review protocol,'' \emph{JBI Evidence Synthesis}, vol.~17, no.~1, pp. 37--42, 2019.

\bibitem{ELEKTRA2022}
C.~Fuster-Barcel\'o, P.~Peris-Lopez, and C.~Camara, ``Elektra: Elektrokardiomatrix application to biometric identification with convolutional neural networks,'' \emph{Neurocomputing (In Press).}, 2022.

\bibitem{ecm}
D.~Li, F.~Tian, S.~Rengifo, G.~Xu, M.~M. Wang, and J.~Borjigin, ``Electrocardiomatrix: A new method for beat-by-beat visualization and inspection of cardiac signals,'' \emph{J Integr Cardiol}, vol.~1, no.~5, pp. 124--128, 2015.

\bibitem{XU2018955}
\BIBentryALTinterwordspacing
G.~Xu, S.~Dodaballapur, T.~Mihaylova, and J.~Borjigin, ``Electrocardiomatrix facilitates qualitative identification of diminished heart rate variability in critically ill patients shortly before cardiac arrest,'' \emph{Journal of Electrocardiology}, vol.~51, no.~6, pp. 955--961, 2018. [Online]. Available: \url{https://www.sciencedirect.com/science/article/pii/S0022073618303480}
\BIBentrySTDinterwordspacing

\bibitem{STROKEAHA025361}
\BIBentryALTinterwordspacing
D.~L. Brown, G.~Xu, A.~M.~B. Krzyske, N.~C. Buhay, M.~Blaha, M.~M. Wang, P.~Farrehi, and J.~Borjigin, ``Electrocardiomatrix facilitates accurate detection of atrial fibrillation in stroke patients,'' \emph{Stroke}, vol.~50, no.~7, pp. 1676--1681, 2019. [Online]. Available: \url{https://www.ahajournals.org/doi/abs/10.1161/STROKEAHA.119.025361}
\BIBentrySTDinterwordspacing

\bibitem{9344257}
R.~{Salinas-Martínez}, J.~{De Bie}, N.~{Marzocchi}, and F.~{Sandberg}, ``Automatic detection of atrial fibrillation using electrocardiomatrix and convolutional neural network,'' in \emph{2020 Computing in Cardiology}, 2020, pp. 1--4.

\bibitem{salinas2021detection}
R.~Salinas-Mart{\'\i}nez, J.~De~Bie, N.~Marzocchi, and F.~Sandberg, ``Detection of brief episodes of atrial fibrillation based on electrocardiomatrix and convolutional neural network,'' \emph{Frontiers in physiology}, vol.~12, 2021.

\bibitem{jaafa2021implementation}
N.~K. Jaafa, B.~Mokaya, S.~M. Savai, A.~Yeung, A.~M. Siika, and M.~Were, ``Implementation of fingerprint technology for unique patient matching and identification at an hiv care and treatment facility in western kenya: Cross-sectional study,'' \emph{Journal of medical Internet research}, vol.~23, no.~12, p. e28958, 2021.

\bibitem{saleem2023face}
S.~Saleem, J.~Shiney, B.~P. Shan, and V.~K. Mishra, ``Face recognition using facial features,'' \emph{Materials Today: Proceedings}, vol.~80, pp. 3857--3862, 2023.

\bibitem{10042627}
F.~Boutros, M.~Klemt, M.~Fang, A.~Kuijper, and N.~Damer, ``Unsupervised face recognition using unlabeled synthetic data,'' in \emph{2023 IEEE 17th International Conference on Automatic Face and Gesture Recognition (FG)}, 2023, pp. 1--8.

\bibitem{khatun2022comparison}
F.~Khatun, R.~Distler, M.~Rahman, B.~O’Donnell, N.~Gachuhi, M.~Alwani, Y.~Wang, A.~Rahman, J.~F. Fr{\o}en, and I.~K. Friberg, ``Comparison of a palm-based biometric solution with a name-based identification system in rural bangladesh,'' \emph{Global health action}, vol.~15, no.~1, p. 2045769, 2022.

\bibitem{amrouni2023palmprint}
N.~Amrouni, A.~Benzaoui, and A.~Zeroual, ``Palmprint recognition: Extensive exploration of databases, methodologies, comparative assessment, and future directions,'' \emph{Applied Sciences}, vol.~14, no.~1, p. 153, 2023.

\bibitem{ostad2019robust}
A.~Ostad-Sharif, D.~Abbasinezhad-Mood, and M.~Nikooghadam, ``A robust and efficient ecc-based mutual authentication and session key generation scheme for healthcare applications,'' \emph{Journal of medical systems}, vol.~43, no.~1, pp. 1--22, 2019.

\bibitem{anne2020feasibility}
N.~Anne, M.~D. Dunbar, F.~Abuna, P.~Simpson, P.~Macharia, B.~Betz, P.~Cherutich, D.~Bukusi, and F.~Carey, ``Feasibility and acceptability of an iris biometric system for unique patient identification in routine hiv services in kenya,'' \emph{International journal of medical informatics}, vol. 133, p. 104006, 2020.

\bibitem{nguyen2024deep}
K.~Nguyen, H.~Proen{\c{c}}a, and F.~Alonso-Fernandez, ``Deep learning for iris recognition: A survey,'' \emph{ACM Computing Surveys}, vol.~56, no.~9, pp. 1--35, 2024.

\bibitem{9035151}
A.~Z. Zahid, I.~H. Mohammed Salih Al-Kharsan, H.~A. Bakarman, M.~F. Ghazi, H.~A. Salman, and F.~N. Hasoon, ``Biometric authentication security system using human dna,'' in \emph{2019 First International Conference of Intelligent Computing and Engineering (ICOICE)}, 2019, pp. 1--7.

\bibitem{jacob2021biometric}
I.~J. Jacob, P.~Betty, P.~E. Darney, S.~Raja, Y.~H. Robinson, and E.~G. Julie, ``Biometric template security using dna codec based transformation,'' \emph{Multimedia Tools and Applications}, vol.~80, pp. 7547--7566, 2021.

\bibitem{aubert2001adoption}
B.~A. Aubert and G.~Hamel, ``Adoption of smart cards in the medical sector:: the canadian experience,'' \emph{Social Science \& Medicine}, vol.~53, no.~7, pp. 879--894, 2001.

\bibitem{sohn2020clinical}
J.~W. Sohn, H.~Kim, S.~B. Park, S.~Lee, J.~I. Monroe, T.~B. Malone, T.~Kinsella, M.~Yao, C.~Kunos, S.~S. Lo \emph{et~al.}, ``Clinical study of using biometrics to identify patient and procedure,'' \emph{Frontiers in Oncology}, vol.~10, p. 586232, 2020.

\bibitem{raghavan2024enhanced}
R.~Raghavan and K.~John~Singh, ``An enhanced and hybrid fingerprint minutiae feature extraction method for identifying and authenticating the patient’s noisy fingerprint,'' \emph{International Journal of System Assurance Engineering and Management}, vol.~15, no.~1, pp. 84--97, 2024.

\bibitem{batool2022causes}
\BIBentryALTinterwordspacing
S.~Batool, H.~Tariq, M.~Shahid, S.~Siddiqui, S.~Batool, and S.~Aman, ``Causes of adermatoglyphia: A hurdle to biometric authentication,'' \emph{Journal of Pakistan Association of Dermatologists}, vol.~32, no.~1, pp. 42--46, 2022. [Online]. Available: \url{https://jpad.com.pk/index.php/jpad/article/view/1805}
\BIBentrySTDinterwordspacing

\bibitem{deneken2022capecitabine}
Z.~Deneken-Hernandez, M.~Cherem-Kibrit, L.~Guti{\'e}rrez-Andrade, G.~Rodr{\'\i}guez-Guti{\'e}rrez, and J.~O. Colmenero-Mercado, ``Capecitabine induced fingerprint loss: Case report and review of the literature,'' \emph{Journal of Oncology Pharmacy Practice}, vol.~28, no.~2, pp. 495--499, 2022.

\bibitem{adjabi2020past}
I.~Adjabi, A.~Ouahabi, A.~Benzaoui, and A.~Taleb-Ahmed, ``Past, present, and future of face recognition: A review,'' \emph{Electronics}, vol.~9, no.~8, p. 1188, 2020.

\bibitem{ali2021classical}
W.~Ali, W.~Tian, S.~U. Din, D.~Iradukunda, and A.~A. Khan, ``Classical and modern face recognition approaches: a complete review,'' \emph{Multimedia tools and applications}, vol.~80, pp. 4825--4880, 2021.

\bibitem{dargie2024identification}
W.~Dargie, S.~Farrokhi, and C.~Poellabauer, ``Identification of persons based on electrocardiogram and motion data,'' \emph{TechRxiv}, 2024.

\bibitem{9231814}
V.~Chandrashekhar, P.~Singh, M.~Paralkar, and O.~K. Tonguz, ``Pulse id: The case for robustness of ecg as a biometric identifier,'' in \emph{2020 IEEE 30th International Workshop on Machine Learning for Signal Processing (MLSP)}, 2020, pp. 1--6.

\bibitem{apandi2022qrs}
Z.~F.~M. Apandi, R.~Ikeura, S.~Hayakawa, and S.~Tsutsumi, ``Qrs detection in electrocardiogram signal of exercise physical activity,'' in \emph{Journal of Physics: Conference Series}, vol. 2319, no.~1.\hskip 1em plus 0.5em minus 0.4em\relax IOP Publishing, 2022, p. 012021.

\bibitem{nsrdb}
A.~Goldberg, ``Physiobank, physiotoolkit, and physionet: Components of a new research resource for complex physiologic signals. circulation [online]. 101 (23), pp. e215–e220,'' Aug 1999.

\bibitem{mitdb}
G.~B. {Moody} and R.~G. {Mark}, ``The impact of the mit-bih arrhythmia database,'' \emph{IEEE Engineering in Medicine and Biology Magazine}, vol.~20, no.~3, pp. 45--50, 2001.

\bibitem{ptbdb}
R.~Bousseljot, D.~Kreiseler, and A.~Schnabel, ``Nutzung der ekg-signaldatenbank cardiodat der ptb über das internet,'' \emph{Biomedizinische Technik / Biomedical Engineering}, vol.~40, pp. 317--318, 1 1995.

\bibitem{gudb}
L.~Howell and B.~Porr, ``High precision ecg database with annotated r peaks, recorded and filmed under realistic conditions,'' 2018.

\bibitem{pantompkins}
T.-W. Shen, W.~Tompkins, and Y.~Hu, ``One-lead ecg for identity verification,'' in \emph{Proceedings of the Second Joint 24th Annual Conference and the Annual Fall Meeting of the Biomedical Engineering Society][Engineering in Medicine and Biology}, vol.~1.\hskip 1em plus 0.5em minus 0.4em\relax IEEE, 2002, pp. 62--63.

\bibitem{MONEDERO2022104536}
\BIBentryALTinterwordspacing
I.~Monedero, ``A novel ecg diagnostic system for the detection of 13 different diseases,'' \emph{Engineering Applications of Artificial Intelligence}, vol. 107, p. 104536, 2022. [Online]. Available: \url{https://www.sciencedirect.com/science/article/pii/S0952197621003845}
\BIBentrySTDinterwordspacing

\bibitem{UBEYLI20081196}
\BIBentryALTinterwordspacing
E.~D. Übeyli, ``Support vector machines for detection of electrocardiographic changes in partial epileptic patients,'' \emph{Engineering Applications of Artificial Intelligence}, vol.~21, no.~8, pp. 1196--1203, 2008. [Online]. Available: \url{https://www.sciencedirect.com/science/article/pii/S0952197608000420}
\BIBentrySTDinterwordspacing

\bibitem{LEE2018S121}
\BIBentryALTinterwordspacing
V.~Lee, G.~Xu, V.~Liu, P.~Farrehi, and J.~Borjigin, ``Accurate detection of atrial fibrillation and atrial flutter using the electrocardiomatrix technique,'' \emph{Journal of Electrocardiology}, vol.~51, no. 6, Supplement, pp. S121--S125, 2018. [Online]. Available: \url{https://www.sciencedirect.com/science/article/pii/S0022073618303406}
\BIBentrySTDinterwordspacing

\bibitem{ecm_2022}
K.~Sharma, M.~Rao, P.~Marwaha, and A.~Kumar, ``Accurate detection of congestive heart failure using electrocardiomatrix technique,'' \emph{Multimedia Tools and Applications}, 04 2022.

\bibitem{STROKEAHA}
D.~L. Brown, G.~Xu, A.~M.~B. Krzyske, N.~C. Buhay, M.~Blaha, M.~M. Wang, P.~Farrehi, and J.~Borjigin, ``Electrocardiomatrix facilitates accurate detection of atrial fibrillation in stroke patients,'' \emph{Stroke}, vol.~50, no.~7, pp. 1676--1681, 2019.

\bibitem{EBRAHIMI2020100033}
\BIBentryALTinterwordspacing
Z.~Ebrahimi, M.~Loni, M.~Daneshtalab, and A.~Gharehbaghi, ``A review on deep learning methods for ecg arrhythmia classification,'' \emph{Expert Systems with Applications: X}, vol.~7, p. 100033, 2020. [Online]. Available: \url{https://www.sciencedirect.com/science/article/pii/S2590188520300123}
\BIBentrySTDinterwordspacing

\bibitem{noran2022towards}
O.~Noran and P.~Bernus, ``Towards an evaluation framework for ubiquitous, self-evolving patient identification solutions in health information systems,'' \emph{Procedia Computer Science}, vol. 196, pp. 550--560, 2022.

\bibitem{li2024biometric}
S.~Li, Y.~Shao, P.~Zan, and H.~Huang, ``Biometric identification based on electrocardiogram using markov transition field and hybrid network,'' in \emph{2024 4th International Conference on Neural Networks, Information and Communication (NNICE)}.\hskip 1em plus 0.5em minus 0.4em\relax IEEE, 2024, pp. 685--688.

\bibitem{10043674}
P.~Melzi, R.~Tolosana, and R.~Vera-Rodriguez, ``Ecg biometric recognition: Review, system proposal, and benchmark evaluation,'' \emph{IEEE Access}, vol.~11, pp. 15\,555--15\,566, 2023.

\bibitem{s20113069}
\BIBentryALTinterwordspacing
B.-H. Kim and J.-Y. Pyun, ``Ecg identification for personal authentication using lstm-based deep recurrent neural networks,'' \emph{Sensors}, vol.~20, no.~11, 2020. [Online]. Available: \url{https://www.mdpi.com/1424-8220/20/11/3069}
\BIBentrySTDinterwordspacing

\bibitem{chee2022electrocardiogram}
K.~J. Chee and D.~A. Ramli, ``Electrocardiogram biometrics using transformer’s self-attention mechanism for sequence pair feature extractor and flexible enrollment scope identification,'' \emph{Sensors}, vol.~22, no.~9, p. 3446, 2022.

\bibitem{el2022wavelet}
I.~El~Boujnouni, H.~Zili, A.~Tali, T.~Tali, and Y.~Laaziz, ``A wavelet-based capsule neural network for ecg biometric identification,'' \emph{Biomedical Signal Processing and Control}, vol.~76, p. 103692, 2022.

\bibitem{prakash2022baed}
A.~J. Prakash, K.~K. Patro, M.~Hammad, R.~Tadeusiewicz, and P.~P{\l}awiak, ``Baed: A secured biometric authentication system using ecg signal based on deep learning techniques,'' \emph{Biocybernetics and Biomedical Engineering}, vol.~42, no.~4, pp. 1081--1093, 2022.

\bibitem{8327810}
J.~Liu, L.~Yin, C.~He, B.~Wen, X.~Hong, and Y.~Li, ``A multiscale autoregressive model-based electrocardiogram identification method,'' \emph{IEEE Access}, vol.~6, pp. 18\,251--18\,263, 2018.

\bibitem{9185990}
J.~R. Pinto and J.~S. Cardoso, ``An end-to-end convolutional neural network for ecg-based biometric authentication,'' in \emph{2019 IEEE 10th International Conference on Biometrics Theory, Applications and Systems (BTAS)}, 2019, pp. 1--8.

\bibitem{DONIDALABATI201978}
\BIBentryALTinterwordspacing
R.~{Donida Labati}, E.~Muñoz, V.~Piuri, R.~Sassi, and F.~Scotti, ``Deep-ecg: Convolutional neural networks for ecg biometric recognition,'' \emph{Pattern Recognition Letters}, vol. 126, pp. 78--85, 2019, robustness, Security and Regulation Aspects in Current Biometric Systems. [Online]. Available: \url{https://www.sciencedirect.com/science/article/pii/S0167865518301077}
\BIBentrySTDinterwordspacing

\bibitem{8695698}
Y.~Chu, H.~Shen, and K.~Huang, ``Ecg authentication method based on parallel multi-scale one-dimensional residual network with center and margin loss,'' \emph{IEEE Access}, vol.~7, pp. 51\,598--51\,607, 2019.

\bibitem{8856916}
P.-L. Hong, J.-Y. Hsiao, C.-H. Chung, Y.-M. Feng, and S.-C. Wu, ``Ecg biometric recognition: Template-free approaches based on deep learning,'' in \emph{2019 41st Annual International Conference of the IEEE Engineering in Medicine and Biology Society (EMBC)}, 2019, pp. 2633--2636.

\bibitem{LI202083}
\BIBentryALTinterwordspacing
Y.~Li, Y.~Pang, K.~Wang, and X.~Li, ``Toward improving ecg biometric identification using cascaded convolutional neural networks,'' \emph{Neurocomputing}, vol. 391, pp. 83--95, 2020. [Online]. Available: \url{https://www.sciencedirect.com/science/article/pii/S0925231220300485}
\BIBentrySTDinterwordspacing

\bibitem{al2024person}
A.~Al-Jibreen, S.~Al-Ahmadi, S.~Islam, and A.~M. Artoli, ``Person identification with arrhythmic ecg signals using deep convolution neural network,'' \emph{Scientific Reports}, vol.~14, no.~1, p. 4431, 2024.

\bibitem{camara2023ecg}
C.~Camara, P.~Peris-Lopez, M.~Safkhani, and N.~Bagheri, ``Ecg identification based on the gramian angular field and tested with individuals in resting and activity states,'' \emph{Sensors}, vol.~23, no.~2, p. 937, 2023.

\end{thebibliography}

\appendices
\appendix 
\section{State of the art regarding the identification of users with ECGs}
This section presents a comparative analysis of state-of-the-art methods for user identification using electrocardiogram (ECG) signals, as shown in Table \ref{tab:comparative-analysis}. A significant distinction of our study is its focus on patient identification, as opposed to general user classification, which is the primary objective in the other referenced studies. This focus on patient identification is crucial for real-life applications in healthcare settings, an aspect not addressed in the majority of the works listed in the table.

The selected studies for this comparative analysis utilize similar datasets for user identification, thereby ensuring a valid comparison. Our method refines and enhances the existing approaches by implementing a straightforward and effective technique that converts ECG signals into matrices, plots them as heatmaps, and then uses a simple two-layer convolutional neural network (CNN) for classification. This simplicity is paramount for practical deployment in healthcare environments.

Several works in the table, such as \cite{li2024biometric} or \cite{10043674}, exhibit lower accuracies. Our proposal improves upon these by emphasizing ease of implementation and robustness. Despite some studies achieving slightly better results, such as those in \cite{s20113069, chee2022electrocardiogram, el2022wavelet, prakash2022baed}, these gains come at the cost of significantly more complex architectures and methodologies for feature extraction and user classification.

For instance, in \cite{8327810}, a non-fiducial feature extraction technique based on autoregressive models was used to identify users from ECG signals. The authors tested this on the PTBDB database, achieving accuracies of 98\% for healthy subjects and 100\% for those with cardiovascular disease (CVD). However, this method requires complex feature extraction processes, and the computational cost is not thoroughly discussed. In contrast, our method achieves similar accuracy on the PTBDB with a more generalized approach, testing over hundreds of images per user.

Other notable works, such as \cite{9185990, DONIDALABATI201978, 8695698, 8856916, LI202083, chee2022electrocardiogram, al2024person}, while effective, do not match the performance of our study in terms of user identification. These methods often employ intricate models which, although accurate, are less practical for straightforward implementation in healthcare systems.

A unique aspect of our research is the investigation of the impact of CVD on individual identification, a crucial factor for healthcare applications. Unlike most of the existing literature, which does not account for the cardiovascular health status of users, our study specifically addresses this gap. We provide valuable insights into how CVD influences identification accuracy, thereby enhancing the reliability of patient identification systems in medical settings.

Overall, our proposed method's distinctive focus on the effects of CVD and varying heartbeat rates on patient identification sets it apart. By considering these physiological factors, we ensure that the results are unbiased and comprehensive. Our method demonstrates consistent and reliable performance with lower error rates and higher accuracy metrics, verified through extensive testing. This makes our approach a significant contribution to the field of user identification in healthcare systems, offering a robust solution for practical applications.

\begin{table*}[t]
\centering
\caption{Comparative analysis of \gls{ecg} identification between our proposal and the state-of-the-art. Only the best result of each study is shown.}\label{tab:comparative-analysis}
\resizebox{0.9\columnwidth}{!}{%
\begin{tabular}{p{1.2cm} c c c p{2cm} c c }\hline
\toprule
\textbf{Proposal} & \textbf{Database} & \textbf{Subjects} & \textbf{Condition} & \textbf{Identification} & \textbf{Metric} & \textbf{Result}\\
 &  &  &  & \textbf{method} &  & (\%) \\\hline
\midrule

\multirow{2}{2cm}{~\cite{8327810}} & \multirow{2}{0.95cm}{PTBDB} & 50 & healthy & \multirow{2}{2cm}{AR model + TM/RF}   & \multirow{2}{4pt}{Acc} & 98.00  \\
 &  & 50 & \gls{cvd} &  &  & 100.00 \\\hline

~\cite{9185990} & PTBDB & 290 & mixed & IT-CNN & EER & 9.06 \\\hline

~\cite{DONIDALABATI201978} & PTBDB & 52 & healthy & CNN & EER & 1.63 \\\hline

\multirow{2}{2cm}{~\cite{8695698}} & PTBDB & 52 & healthy & 
\multirow{2}{2cm}{CNN + Euclidian dist. }  & \multirow{2}{2pt}{EER} & 0.59  \\
 & MITDB & 47 & mixed &  &  & 4.74 \\\hline

~\cite{8856916} & PTBDB & 200 & mixed & pre-trained CNN & IR & 98.1 \\\hline

\multirow{2}{2cm}{~\cite{s20113069}} & MITDB & 48 & mixed & \multirow{2}{2cm}{biLSTM}  & \multirow{2}{2pt}{Acc} & 99.80  \\
 & NSRDB & 18 & healthy &  &  & 100.00 \\\hline

\cite{LI202083} & NSRDB & 18 & healthy & Cascaded-CNN & IR & 91.40 \\\hline

\multirow{2}{2cm}{~\cite{chee2022electrocardiogram}} & PTBDB & 52 & healthy & \multirow{2}{2cm}{BERT}  & \multirow{2}{2pt}{Acc} & 98.10  \\
 & NSRDB & 18 & healthy &  &  & 99.91 \\\hline

\multirow{3}{2cm}{~\cite{el2022wavelet}} & PTBDB & 52 & healthy & \multirow{3}{2cm}{Scalogram + CapsNet}  & \multirow{3}{2pt}{Acc} & 98.8  \\
 & NSRDB & 18 & healthy &  &  & 100.00 \\
 & MITDB & 48 & mixed &  &  & 98.20 \\\hline

\cite{prakash2022baed} & {PTBDB} & {290} & {mixed} & {CNN+LSTM} & {Acc} & {99.62} \\\hline
 
\multirow{2}{2cm}{~\cite{camara2023ecg}} & GUDB & 25 & healthy & \multirow{2}{2cm}{GAF + VGG19}  & \multirow{2}{2pt}{Acc} & 91.60  \\
 & NSRDB & 18 & healthy &  &  & 91.00 \\\hline

\cite{10043674} & {PTBDB} & {113} & {mixed} & {AE + CNN} & {Acc} & {96.46} \\\hline

\cite{al2024person} & MITDB & 47 & mixed & DWSC & Acc & 99.58 \\\hline

\cite{li2024biometric} & GUDB & 25 & healthy & Hybrid Network & Acc & 92.73 \\\hline

\multirow{12}{2cm}{Our Proposal} & \multirow{3}{2pt}{NSRDB} & \multirow{3}{2pt}{18} & \multirow{3}{2pt}{healthy} &\multirow{12}{5cm}{ECM + CNN} & Acc & 99.84 \\
& & & & & IR & 99.98 \\
& & & & & EER & 0.08\\ \cline{2-4} \cline{6-7}

& \multirow{3}{2pt}{MITDB} & \multirow{3}{2pt}{47} & \multirow{3}{2pt}{healthy/ arrhythmia} & & Acc & 97.89 \\
& & & & & IR & 99.91 \\
& & & & & EER & 1.075 \\ \cline{2-4} \cline{6-7}

& \multirow{3}{2pt}{PTBDB} & \multirow{3}{2pt}{162} & \multirow{3}{2pt}{CVD} & & Acc & 97.09 \\
& & & & & IR & 99.96 \\
& & & & & EER & 1.465\\ \cline{2-4} \cline{6-7}

& \multirow{3}{2pt}{GUDB} & \multirow{3}{2pt}{25} & \multirow{3}{2pt}{healthy sitting} & & Acc & 99.19 \\
& & & & & IR & 99.93 \\
& & & & & EER & 0.42 \\

\bottomrule
\end{tabular}%
}
\begin{quote}
\scriptsize
\centering
$^{\dagger}$Notation: Acc (Accuracy); EER (Equal Error Rate); IR (Identification Rate); Healthy (Users with not significant \gls{cvd}); CVD (Patients with \gls{cvd}); mixed (healthy users and patients with \gls{cvd}). 
\end{quote}
\end{table*}

\section*{Data availability}
All utilised datasets are either accessible online (\cite{nsrdb, mitdb, ptbdb}) or can be obtained upon request (\cite{gudb}) to guarantee the replicability of all the experiments. 

\section*{Credit authorship contribution statement}
Caterina Fuster-Barceló: Conceptualisation,  Methodology, Experimentation,  Supervision, Writing - original draft, Writing - review \& editing, Validation.  Pedro Peris-Lopez: Conceptualisation, Methodology, Supervision, Funding,  Writing - review \& editing, Validation. Carmen Camara: Conceptualisation, Methodology,  Supervision, Funding, Writing - review \& editing, Validation. 

\section*{Declaration of competing interest} 
The authors declare that they have no known competing financial interests or personal relationships that could have influenced the work reported in this article.

\section*{Acknowledgment}

This work was supported by grants TED2021-131681B-I00 (CIOMET) and PID2022-140126OB-I00 (CYCAD) from the Spanish Ministry of Science, Innovation and Universities, as well as by the INCIBE under the project VITAL-IoT in the context of the funds from the Recovery, Transformation, and Resilience Plan, financed by the European Union (Next Generation).

The authors gratefully acknowledge the computer resources at Artemisa, funded by the European Union ERDF and Comunitat Valenciana as well as the technical support provided by the Instituto de Fisica Corpuscular, IFIC (CSIC-UV).

This article was enhanced for grammar and clarity using ChatGPT, an AI language model developed by OpenAI.
\end{document}